\documentclass[onefignum,onetabnum]{siamonline190516}

\usepackage{lipsum}
\usepackage{amsfonts, amssymb, mathrsfs}
\usepackage{graphicx}
\usepackage{epstopdf}
\usepackage{algorithmic}
\usepackage{csquotes}   
\usepackage{tikz, tikz-3dplot}
\usetikzlibrary{3d}
\usetikzlibrary{arrows, arrows.meta,calc,decorations.markings,math,decorations.pathmorphing, positioning, shapes,backgrounds,fit}
\usepackage[most]{tcolorbox} 
\usepackage{relsize}  
\usepackage{tabularx, makecell, hhline, bigints, multirow} 
\allowdisplaybreaks

\graphicspath{{images/}}
\protected\def\tikz@nonactivecolon{\ifmmode\mathrel{\mathop\ordinarycolon}\else:\fi} 

\ifpdf
  \DeclareGraphicsExtensions{.eps,.pdf,.png,.jpg}
\else
  \DeclareGraphicsExtensions{.eps}
\fi

\usepackage{enumitem}
\setlist[enumerate]{leftmargin=.5in}
\setlist[itemize]{leftmargin=.5in}

\newsiamremark{remark}{Remark}
\newsiamremark{hypothesis}{Hypothesis}
\crefname{hypothesis}{Hypothesis}{Hypotheses}
\newsiamthm{claim}{Claim}

\headers{Deep Neural Networks and PIDE discretizations}{B.Bohn, M. Griebel, and D. Kannan}

\title{Deep Neural Networks and PIDE discretizations}

\author{Bastian Bohn\footnote[1]{Fraunhofer Center for Machine Learning, Schloss Birlinghoven, 53754 Sankt Augustin, Germany (\email{bastian.bohn@scai.fraunhofer.de}).} \footnote[2]{Fraunhofer Institute for Algorithms and Scientific Computing SCAI, Schloss Birlinghoven, 53754 Sankt Augustin, Germany.}
\and Michael Griebel\footnotemark[2] \footnote[3]{Institute for Numerical Simulation, University of Bonn, Friedrich-Hirzebruch-Allee 7, 53115 Bonn, Germany (\email{griebel@ins.uni-bonn.de}, \email{kannan@ins.uni-bonn.de}).} \and Dinesh Kannan\footnotemark[3]}

\usepackage{amsopn}

\DeclareMathOperator{\var}{Var}

\ifpdf
\hypersetup{
  pdftitle={Deep Neural Networks and PIDE discretizations},
  pdfauthor={B. Bohn, M. Griebel, and D. Kannan}
}
\fi

\begin{document}

\maketitle

\begin{abstract}
In this paper, we propose neural networks that tackle the problems of stability and field-of-view of a Convolutional Neural Network (CNN).  As an alternative to increasing the network's depth or width to improve performance, we propose integral-based spatially nonlocal operators which are related to global weighted Laplacian, fractional
Laplacian and inverse fractional Laplacian operators that arise in several problems
in the physical sciences. The forward propagation of such networks is inspired by partial integro-differential equations (PIDEs). We test the effectiveness of the proposed neural architectures on benchmark image classification datasets and semantic segmentation tasks in autonomous driving. Moreover, we investigate the extra computational costs of these dense operators and the stability of forward propagation of the proposed neural networks. 
\end{abstract}



\section{Introduction}
We consider the tasks of image classification and segmentation which can be seen as a data fitting problem in a high-dimensional space. Image classification involves assigning each element of a dataset to one of the many classes or categories of images. It forms the basis for several other computer vision tasks such as detection, segmentation and localization. Let each $C$-channel image be of dimension $\mathbb{R}^{H\times W \times C}$, with each image belonging to one of the $k$ classes. Then the objective is to find a classifier map $\mathcal{J}: \mathbb{R}^{H\times W \times C}\mapsto \mathbb{R}^k$ such that, for a given $\mathbf x \in \mathbb{R}^{H\times W \times C}$, we have $\mathcal{J}(\mathbf x) \approx p(\mathbf x) \in \mathbb{R}^k$, where $p(\mathbf x)$ represents the probabilities of $\mathbf x$ belonging to one of the $k$ classes. The task of semantic segmentation is similar and can be seen as a multi-dimensional extension of the image classification task. Here, each pixel of each image is supposed to be assigned to one of the many classes (car, bus, etc.), i.e. each pixel gets a meaning so that, in the end, individual objects on an image can be distinguished from one another (see Section \ref{sec:segment}).

Deep Neural Networks (DNNs) can be used to train machines to learn from such examples, to recognize patterns in the data effectively and to make predictions on unseen examples \cite{goodfellow2016}. DNNs have found several applications in the last decade, such as natural language processing \cite{deepuse1} and solving partial differential equations (PDEs) \cite{pdeuse}, just to name a few. Convolutional Neural Networks (CNNs) are a special class of neural networks that make use of spatial correlations between features by introducing a series of spatial convolution operators into the network with compactly supported stencils/kernels and point-wise nonlinearities, where the trainable kernel weights capture hierarchical patterns. CNNs increase the computational efficiency of the network due to the sparse connections between the features and due to the reduction in the number of weights via parameter sharing. Moreover, in \cite{universal}, it is argued that CNNs can approximate any continuous function to an arbitrary accuracy if the depth of the neural network is large enough.

There are several hurdles when it comes to designing and training a neural network. Firstly, there is the well-known issue of exploding and vanishing gradients \cite{expvan}. Also, in CNNs, the convolution operation is local, and, as a consequence, each layer draws inferences based on the information that is present in a small spatial neighborhood of the image (receptive field) and therefore lacks a holistic view of the feature maps. This is known as the \emph{field-of-view} problem \cite{fov}. To make sure that the output layer has a large receptive field and no information of the image is left out while making the inference, modern architectures tend to go deep \cite{he2016, sim2014}. By stacking more convolutional and pooling layers, the receptive field is increased, and large parts of the image end up having an indirect influence on the output layer. This is helpful for datasets with long-range spatial dependencies. However, the number of trainable weights increases rapidly with depth, and this imposes challenges when using mediocre computational hardware with memory constraints. Thus, the dilemma is whether one should have a very deep network that has a better performance but makes the training harder for any optimizer or to have shallower networks that are easier to train but perform relatively worse.

To this end, we propose neural networks that tackle the field-of-view problem of a CNN. Instead of increasing the network's depth or width to improve a neural network's performance, the introduction of nonlocal operators is advocated. These integral-based spatially nonlocal operators address the issue of field-of-view and reduce the need for deeper networks if one needs better performance. The associated forward propagation is inspired by partial integro-differential equations (PIDEs). Specifically, a corresponding layer that is inspired by a specific PIDE system is added to a ResNet network of consecutive Hamiltonian blocks. We examine the effectiveness of the proposed network designs for several image classification benchmark datasets. Our networks need lesser depth than the conventional CNNs to obtain the same accuracy and expressibility while maintaining the numerical stability of the network. Additionally, this paper demonstrates a real-world application, namely our networks are used for semantic segmentation tasks in autonomous driving. 

This paper is organized as follows. In Section 2, a general introduction to Residual Neural Networks (ResNets), its mathematical formulation and its link with stable dynamical systems is reviewed. Section 3 introduces the PIDE-inspired nonlocal operators, and a few inherent properties of such operators are explored and reviewed. These nonlocal operators are then discretized in Section 4, and several strategies to save computational costs are proposed. In this section, also the implementation details for the nonlocal operators and the proposed CNNs are discussed. In Section 5, numerical experiments related to image classification and semantic segmentation are conducted, and their effectiveness is examined. Section 6 deals with the forward propagation stability of the proposed CNNs and the computational cost associated with them. The concluding Section 7 summarizes the results.

\section{Mathematical formulation}
In this section, we briefly formulate the classification task as a parameter estimation problem. We also introduce the building blocks of the basic neural architecture that will be used in the subsequent sections of this paper. For a detailed summary of deep learning, see \cite{goodfellow2016, introdeep}.

Let $\mathbf Y_0$ be the input to the neural network. The forward propagation of a Residual Neural Network (ResNet) with $N$ layers is given by
\begin{equation}\label{eq:resnet}
\mathbf Y_{j+1} = \mathbf Y_j + \mathcal F(\mathbf Y_j,\mathbf K_j) \quad \text{for } \: j = 0,\hdots,N-1,
\end{equation} where $\mathbf Y_j$ are the feature values in the $j$-th hidden layer, $\mathbf K_j$ are the learnable (convolutional and batch normalization) weights in the $j$-th hidden layer, and $\mathcal F$ is the residual module that can come in various forms, for example, $\mathcal F(\mathbf Y_j, \mathbf K_j) = \sigma(\mathcal B(\mathbf K_{j,2} \, \sigma (\mathcal B(\mathbf K_{j,1} \mathbf Y_j + b_j))))$ \cite{he2016}, where $\mathbf K_j = \{\mathbf K_{j,1}, \mathbf K_{j,2}\}$, i.e. the residual module has two convolutional layers. $\mathcal{B}$ represents the batch normalization (BN) layer \cite{ioffe2015} and $\sigma(\cdot)$ is the activation function applied component-wise. For our purposes, we will stick with the the rectifier or ReLU activation function, $\sigma(x) = \max(0,x)$. $\mathbf Y_N$ is the final layer of the neural network. For multi-class classification tasks, this final layer is passed through a softmax classifier, $S:\mathbb{R}^k \mapsto \mathbb{R}^k$, $[S(x)]_i:= \frac{e^{x_i}}{\sum_{j=1}^k e^{x_j}}$. It returns the necessary probability distribution consisting of $k$ probabilities that are proportional to the entries of the input vector $x$ and are actually the different probabilities of the image/pixel belonging to each of the $k$ classes. 

Essentially, in each step of the iteration, the residual module tries to learn the update made to $\mathbf Y_j$. Note that the residual module depends not just on layer weights $\mathbf K_j$, but also on layer biases $b_j$. For simplicity, however, we write $\mathcal F(\mathbf Y_j,\mathbf K_j)$ instead of $\mathcal F(\mathbf Y_j,\mathbf K_j, b_j)$. In the original paper \cite{he2016}, this residual block followed the sequence of convolution-BN-ReLU. In a follow-up paper \cite{he2016b}, He, Zhang, Ren et al. suggest another sequence for computations, namely BN-ReLU-convolution. Over the last couple of years, ResNet has had a lot of success in several areas, such as object detection \cite{he2017, he2016} and semantic segmentation \cite{segment}. ResNets have made it possible to train up to hundreds of layers and still achieve compelling performance, something which was not possible before. Szegedy, Ioffe, Vanhoucke et al. showed in \cite{incep3} that residual links in fact speed up the convergence of very deep networks.

Let the output of the softmax layer be called $\mathbf Y_{N+1}$, with $\mathbf W$ representing the weights and biases of this extra layer. Let the collection of weights and biases of the other layers be represented as $\mathbf K= \{\mathbf K_j\}_{j=1}^{j=N}$, $b= \{b_j\}_{j=1}^{j=N}$, respectively. Then the learning problem can be cast as a high-dimensional, non-convex optimization problem
\begin{gather}\label{eq:invprob}
\begin{aligned}
&\min_{\mathbf K, \mathbf W, b} & &\frac{1}{2}S(\mathbf Y_{N+1}, \mathbf C) + R(\mathbf K, \mathbf W, b)\\
\text{such }&\text{that} &  &\mathbf Y_{j+1} = \mathbf Y_j + \mathcal F(\mathbf Y_j,\mathbf K_j) \quad \text{for } \: j = 0,\hdots,N-1,\\
\end{aligned} 
\end{gather} where $S$ is the loss function that is convex in the first argument and measures the closeness of the prediction to the ground truth. Usually, $S$ is assumed to be the least-squares function and cross-entropy loss function for regression and classification tasks, respectively \cite{goodfellow2016}. $\mathbf C$ is the label or the ground truth for the example $\mathbf Y_0$ that is propagated through the network. $R$ is a convex regularizer that penalizes large weights that are undesirable (see Section \ref{sec:reg}).

The forward propagation of a ResNet can be interpreted as a discretization of an underlying nonlinear ordinary differential equation (ODE), and in special cases, as a discretization of an underlying partial differential equation (PDE). We add a step size $h$ to the equation of the forward propagation of ResNets \cite{weinan2017, haber2017}, i.e. 
\begin{equation}\label{eq:euler}
\mathbf Y_{j+1} = \mathbf Y_j + h\mathcal F(\mathbf Y_j,\mathbf K_j) \quad \text{for } \: j = 0,\hdots,N-1.
\end{equation}This equation can be interpreted as a forward Euler discretization of the initial value problem \begin{equation}\label{eq:ode}
\dot{\mathbf Y} (t) =  \mathcal F (\mathbf Y(t),\mathbf K(t)), \quad \mathbf Y(0) = \mathbf Y_0,
\end{equation} with the input layer being $\mathbf Y(0)$ and the output layer values $\mathbf Y(T)$ being the solution to the initial value problem at time $T$, with $T= Nh$, $T$ and $N$ being related to the depth of the network. Thus, each layer of the network can be seen as the discretization of the continuous solution at discrete time points in $[0,T]$, and the problem of learning the network weights $\mathbf K_j$ can be seen as a parameter estimation problem or an optimal control problem that is governed by equation \eqref{eq:ode}. Moreover, in \cite{giantleap}, Zhang, Han, Wynter et al. claim that the presence of the step size $h$ in ResNet-like networks stabilizes the gradients and allows larger learning rates.

With the introduction of the dummy time variable $t$, we have a time derivative in the forward propagation. In special cases, such as image classification or segmentation, one mainly has spatial convolution weights $\mathbf K_j$ for each layer, instead of dense weight matrices. In such cases, the convolution weights can be expressed as a linear combination of spatial differential operators or as a coupled system of partial differential operators in case of multi-channel convolutions \cite{pde}. As a result, in the case of CNNs, equation \eqref{eq:resnet} is a forward Euler discretization (in the time domain) of an underlying partial differential equation due to the inherent presence of spatial derivatives in the residual function $\mathcal F$. The CNN layer weights determine the order, type and stability of the underlying PDE.\label{re:pderemark}

The stability of the forward propagation of a neural network is of prime importance. The output of a neural network or the solution to the underlying PDE/ODE is sensitive to perturbations of the input data that are not perceived by the human eye, which makes a network vulnerable to adversarial attacks \cite{adv1}. Similar images might propagate through the network and yield vastly different outputs and predictions \cite{perturb2}, and therefore the network generalizes poorly on test data. Also, such instabilities make the network harder to train via backpropagation. Chang, Meng, Haber et al. suggest several stable neural architectures in \cite{rev}, which are inspired by stable and well-posed solutions of the underlying PDE/ODE of the neural network \cite{rev, haber2017}. Among all the three ODE-based networks suggested in \cite{rev}, the so-called \textit{Hamiltonian network} performs the best on standard benchmarks, possibly due to its two-layer network, where each \textit{Hamiltonian block} has two convolution and two transposed convolution operations. Hamiltonian systems are known to conserve energy, and hence, the forward propagation, in this case, is expected to only moderately amplify or damp the features, as it is propagated down the network. By introducing an extra augmented variable $\mathbf Z(t)$ and two sets of convolution filters $\mathbf K_1(t), \mathbf K_2(t)$, the forward propagation can be interpreted as a Hamiltonian system, where we have a system of PDEs\begin{gather}\label{eq:ham}\begin{aligned}\dot{\mathbf Y} (t) &=  \mathbf K_1^T(t)\,\sigma(\mathbf K_1(t)\mathbf Z(t)+b_1(t)),\\
\dot{\mathbf Z} (t) &=  -\mathbf K_2^T(t)\,\sigma(\mathbf K_2(t)\mathbf Y(t)+b_2(t)).\end{aligned}\end{gather}Here, $\mathbf Y(t)$ and $\mathbf Z(t)$ are channel-wise partitions of the input features, and for our purposes of image classification and segmentation, $\mathbf K_i(t)$ is the convolution operator, and $\mathbf K_i^T(t)$ is its transpose, $i=1,2$. The discretization of equation \eqref{eq:ham} is done using the Verlet integration since such symplectic methods capture the long time features of Hamiltonian systems really well \cite{ascher2}. With the Verlet scheme, we arrive at the following pair of coupled equations shown below. This computation is placed in a Hamiltonian block (see \cite{rev}).
\begin{gather}\label{eq:verlet}
\begin{aligned}
\mathbf Y_{j+1} &= \mathbf Y_j + h \,  \mathbf K_{j1}^T\,\sigma(\mathbf K_{j1}\mathbf Z_j+b_{j1}),\\
\mathbf Z_{j+1} &= \mathbf Z_j - h \,  \mathbf K_{j2}^T\,\sigma(\mathbf K_{j2}\mathbf Y_{j+1}+b_{j2}).
\end{aligned}
\end{gather} 
\ifx
$\mathbf X_j$ is assumed to be the input to the Hamiltonian block, $\mathbf Y_j, \mathbf Z_j$ are channel-wise partitions of $\mathbf X_j$, and $\mathbf X_{j+1}$ is the output of the block that is obtained by the channel-wise concatenation of $\mathbf Y_{j+1}$ and $\mathbf Z_{j+1}$.

\begin{figure}[h!]
	\centering
\begin{tikzpicture}[scale=0.66]

\node[black] at (0.3, 1.2) {\large $\mathbf X_j$};
\draw[fill=cyan, opacity = 0.3, draw=black] (0.8,0.7) -- (1.8,0.7) -- (1.8,1.7) -- (0.8,1.7) -- (0.8,0.7);		
\draw[fill=red,opacity=0.3,draw=black] (0.9,0.6) -- (1.9,0.6) -- (1.9,1.6) -- (0.9,1.6) -- (0.9,0.6);
\draw[fill=red!50!yellow,opacity=0.3,draw=black] (1,0.5) -- (2,0.5) -- (2,1.5) -- (1,1.5) -- (1,0.5);
\draw[fill=yellow,opacity=0.3,draw=black] (1.1,0.4) -- (2.1,0.4) -- (2.1,1.4) -- (1.1,1.4) -- (1.1,0.4);
\draw[fill=green,opacity=0.3,draw=black] (1.2,0.3) -- (2.2,0.3) -- (2.2,1.3) -- (1.2,1.3) -- (1.2,0.3);
\draw[fill=blue,opacity=0.3,draw=black] (1.3,0.2) -- (2.3,0.2) -- (2.3,1.2) -- (1.3,1.2) -- (1.3,0.2);
\draw[fill=blue!50!red,opacity=0.3,draw=black] (1.4,0.1) -- (2.4,0.1) -- (2.4,1.1) -- (1.4,1.1) -- (1.4,0.1);
\draw[fill=blue!50!magenta,opacity=0.3,draw=black] (1.5,0) -- (2.5,0) -- (2.5,1) -- (1.5,1) -- (1.5,0);

\node[inner sep=0pt] at (2,0.5){\includegraphics[width=.04\textwidth]{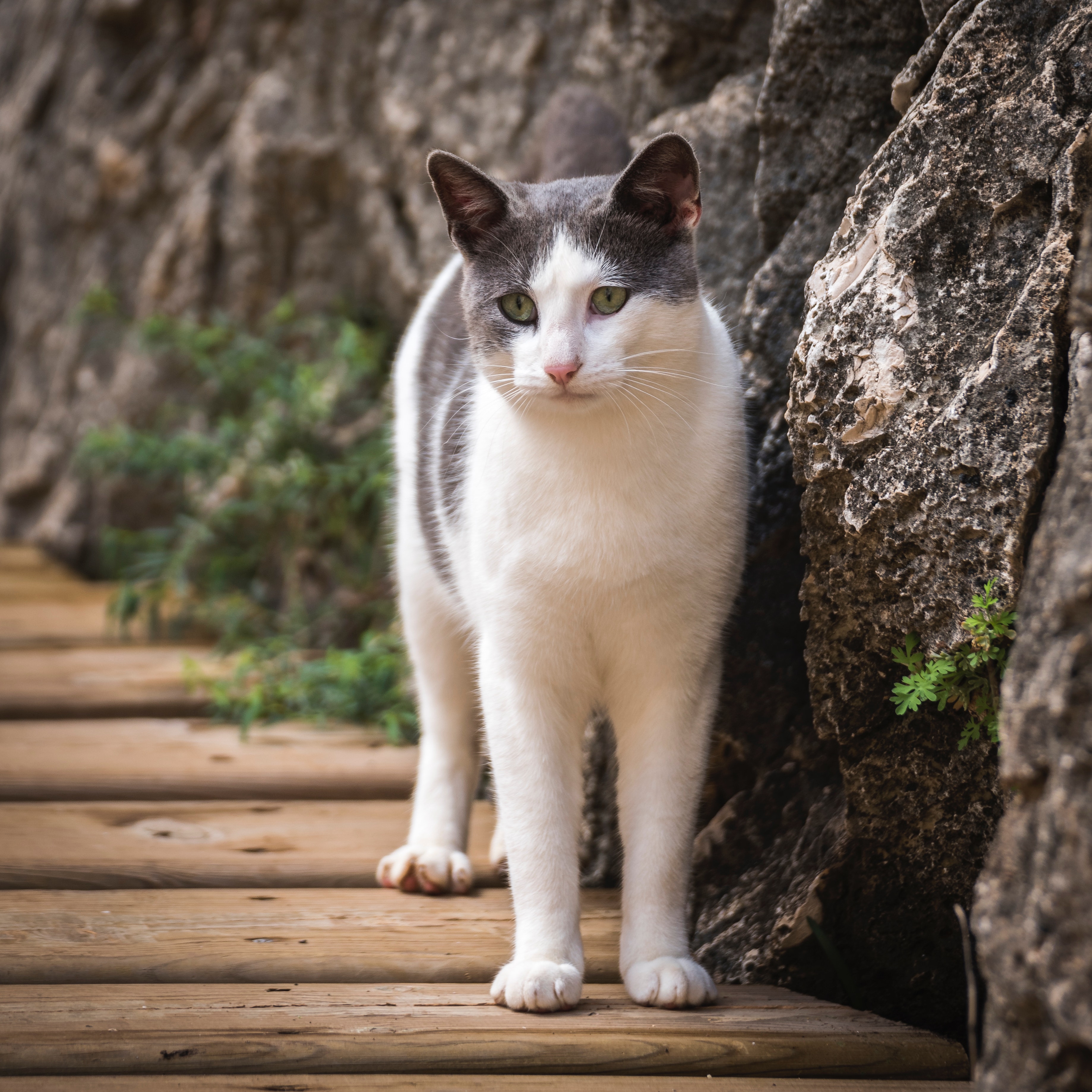}};
\draw[->, -{Latex[scale=0.7]}, black, ultra thick] (2.2,-0.1) --  (4.2,-2.1);	
\draw[->, -{Latex[scale=0.7]}, black, ultra thick] (2.2,1.4) --  (3.7,3.4);	

\node[black] at (3.2, -2.7) {\large $\mathbf Y_j$};
\draw[fill=green,opacity=0.3,draw=black] (3.7,-2.2) -- (4.7,-2.2) -- (4.7,-3.2) -- (3.7,-3.2) -- (3.7,-2.2);
\draw[fill=blue,opacity=0.3,draw=black] (3.8,-2.3) -- (4.8,-2.3) -- (4.8,-3.3) -- (3.8,-3.3) -- (3.8,-2.3);
\draw[fill=cyan!50!red,opacity=0.3,draw=black] (3.9,-2.4) -- (4.9,-2.4) -- (4.9,-3.4) -- (3.9,-3.4) -- (3.9,-2.4);
\draw[fill=blue!50!magenta,opacity=0.3,draw=black] (4,-2.5) -- (5,-2.5) -- (5,-3.5) -- (4,-3.5) -- (4,-2.5);
\node[inner sep=0pt] at (4.5,-3){\includegraphics[width=.04\textwidth]{cat.jpg}};

\node[black] at (3.2, 4.1) {\large $\mathbf Z_j$};
\draw[fill=cyan, opacity = 0.3, draw=black] (3.7,3.6) -- (4.7,3.6) -- (4.7,4.6) -- (3.7,4.6) -- (3.7,3.6);
\draw[fill=red,opacity=0.3,draw=black] (3.8,3.5) -- (4.8,3.5) -- (4.8,4.5) -- (3.8,4.5) -- (3.8,3.5);
\draw[fill=red!50!magenta,opacity=0.3,draw=black] (3.9,3.4) -- (4.9,3.4) -- (4.9,4.4) -- (3.9,4.4) -- (3.9,3.4);
\draw[fill=yellow,opacity=0.3,draw=black] (4,3.3) -- (5,3.3) -- (5,4.3) -- (4,4.3) -- (4,3.3);
\node[inner sep=0pt] at (4.5,3.8){\includegraphics[width=.04\textwidth]{cat.jpg}};

\draw[->, -{Latex[scale=0.7]}, black, ultra thick] (5.1,3.8) --  (10.85,3.8);

\draw[->, -{Latex[scale=0.7]}, black, ultra thick] (6.6,3.8) --  (6.6,2.8);
\draw[thin, draw=black, fill=magenta, fill opacity=0.2, rounded corners=2] (4.7,2.8) rectangle (8.5, 2);
\node [anchor = center, black] at (6.6,2.4) {\scriptsize\bf Conv. $\mathbf K_1$+Bias};

\draw[->, -{Latex[scale=0.7]}, black, ultra thick] (6.6,2) --  (6.6,1);
\draw[thin, draw=black, fill=orange, fill opacity=0.2, rounded corners=2] (4.7,1) rectangle (8.5, 0.2);
\node [anchor = center, black] at (6.6,0.6) {\footnotesize\bf BN + ReLU};

\draw[->, -{Latex[scale=0.7]}, black, ultra thick] (6.6,0.2) --  (6.6,-0.8);
\draw[thin, draw=black, fill=magenta, fill opacity=0.2, rounded corners=2] (4.7,-0.8) rectangle (8.5, -1.6);
\node [anchor = center, black] at (6.6,-1.2) {\footnotesize\bf Conv. $\mathbf K_1^{T}$};

\draw[->, -{Latex[scale=0.7]}, black, ultra thick] (5.1,-3) --  (8.55,-3);
\draw[black, thick, fill=green!50!yellow, fill opacity=0.4] (8.9, -3) circle (11pt);
\node [anchor = center, black] at (8.9,-3) {\LARGE $+$};
\draw[->, -{Latex[scale=0.7]}, black, ultra thick] (9.3,-3) --  (12.7,-3);

\draw[->, -{Latex[scale=0.7]}, black, ultra thick] (6.6,-1.6) to [out=-50,in=160] node[black, midway, above, rotate=-20] {$h$} (8.6,-2.8);

\draw[->, -{Latex[scale=0.7]}, black, ultra thick] (9.25,-2.8) to [out=10,in=-130] (11.2,-1.6);
\draw[thin, draw=black, fill=magenta, fill opacity=0.2, rounded corners=2] (9.3,-0.8) rectangle (13.1, -1.6);
\node [anchor = center, black] at (11.2,-1.2) {\scriptsize\bf Conv. $\mathbf K_2$+Bias};	
\draw[->, -{Latex[scale=0.7]}, black, ultra thick] (11.2,-0.8) --  (11.2,0.2);

\draw[thin, draw=black, fill=orange, fill opacity=0.2, rounded corners=2] (9.3,1) rectangle (13.1, 0.2);
\node [anchor = center, black] at (11.2,0.6) {\footnotesize\bf BN + ReLU};
\draw[->, -{Latex[scale=0.7]}, black, ultra thick] (11.2,1) --  (11.2,2);

\draw[thin, draw=black, fill=magenta, fill opacity=0.2, rounded corners=2] (9.3,2.8) rectangle (13.1, 2);
\node [anchor = center, black] at (11.2,2.4) {\footnotesize\bf Conv. $\mathbf K_2^{T}$};
\draw[->, -{Latex[scale=0.7]}, black, ultra thick] (11.2,2.8) --  (11.2,3.4) node[black, midway, left, xshift = -0.05cm] {$h$};

\draw[black, thick, fill=red!50!blue, fill opacity=0.4] (11.2, 3.8) circle (11pt);
\node [anchor = center, black] at (11.2,3.8) {\LARGE $-$};

\draw[->, -{Latex[scale=0.7]}, black, ultra thick] (11.6,3.8) --  (12.7,3.8);

\node[black] at (14.8, 4.1) {\large $\mathbf Z_{j+1}$};
\draw[fill=red!50!yellow,opacity=0.3,draw=black] (12.8,3.6) -- (13.8,3.6) -- (13.8,4.6) -- (12.8,4.6) -- (12.8,3.6);
\draw[fill=yellow!50!magenta,opacity=0.3,draw=black] (12.9,3.5) -- (13.9,3.5) -- (13.9,4.5) -- (12.9,4.5) -- (12.9,3.5);
\draw[fill=cyan!50!yellow,opacity=0.3,draw=black] (13,3.4) -- (14,3.4) -- (14,4.4) -- (13,4.4) -- (13,3.4);
\draw[fill=red!50!magenta,opacity=0.3,draw=black] (13.1, 3.3) -- (14.1,3.3) -- (14.1,4.3) -- (13.1,4.3) -- (13.1,3.3);
\node[inner sep=0pt] at (13.6,3.8){\includegraphics[width=.04\textwidth]{cat.jpg}};

\node[black] at (14.8, -3.3) {\large $\mathbf Y_{j+1}$};
\draw[fill=red!50!green,opacity=0.3,draw=black] (12.8, -3.5) -- (13.8,-3.5) -- (13.8,-2.5) -- (12.8,-2.5) -- (12.8,-3.5);
\draw[fill=magenta!50!red,opacity=0.3,draw=black] (12.9, -3.6) -- (13.9,-3.6) -- (13.9,-2.6) -- (12.9,-2.6) -- (12.9,-3.6);
\draw[fill=cyan!50!yellow,opacity=0.3,draw=black] (13,-3.7) -- (14,-3.7) -- (14,-2.7) -- (13,-2.7) -- (13,-3.7);
\draw[fill=cyan!50!magenta,opacity=0.3,draw=black] (13.1,-3.8) -- (14.1,-3.8) -- (14.1,-2.8) -- (13.1,-2.8) -- (13.1,-3.8);
\node[inner sep=0pt] at (13.6,-3.3){\includegraphics[width=.04\textwidth]{cat.jpg}};

\draw[->, -{Latex[scale=0.7]}, black, ultra thick] (14,-2.5) --  (15.6,-0.4);	
\draw[->, -{Latex[scale=0.7]}, black, ultra thick] (13.6,3.2) --  (15.6,1.4);	

\draw[fill=yellow,opacity=0.3,draw=black] (15,0.3) -- (16,0.3) -- (16,1.3) -- (15,1.3) -- (15,0.3);
\draw[fill=red,opacity=0.3,draw=black] (15.1,0.2) -- (16.1,0.2) -- (16.1,1.2) -- (15.1,1.2) -- (15.1,0.2);
\draw[fill=blue,opacity=0.3,draw=black] (15.2,0.1) -- (16.2,0.1) -- (16.2,1.1) -- (15.2,1.1) -- (15.2,0.1);
\draw[fill=green,opacity=0.3,draw=black] (15.3,0) -- (16.3,0) -- (16.3,1) -- (15.3,1) -- (15.3,0);
\draw[fill=blue!70!red,opacity=0.3,draw=black] (15.4,-0.1) -- (16.4,-0.1) -- (16.4,0.9) -- (15.4,0.9) -- (15.4,-0.1);
\draw[fill=yellow!70!green,opacity=0.3,draw=black] (15.5,-0.2) -- (16.5,-0.2) -- (16.5,0.8) -- (15.5,0.8) -- (15.5,-0.2);
\draw[fill=cyan!70!magenta,opacity=0.3,draw=black] (15.6,-0.3) -- (16.6,-0.3) -- (16.6,0.7) -- (15.6,0.7) -- (15.6,-0.3);
\draw[fill=magenta,opacity=0.3,draw=black] (15.7,-0.4) -- (16.7,-0.4) -- (16.7,0.6) -- (15.7,0.6) -- (15.7,-0.4);
\node[inner sep=0pt] at (16.2,0.1){\includegraphics[width=.04\textwidth]{cat.jpg}};
\node[black] at (17.3, 0.8) {\large $\mathbf X_{j+1}$};

	\end{tikzpicture}
	\caption{The Hamiltonian block with channel-wise partition and concatenation.}
	\label{fig:ham}
\end{figure}
\fi
\section{Nonlocal and pseudo-differential operators for Deep Learning}
In a Convolutional Neural Network, the pooling or subsampling of the image and the convolution operations allow far-away pixels in an image to communicate with each other, thereby exploiting long-range correlations in an image and improving the predictions made by the network. At the same time, convolution is a local operation, and therefore each convolutional layer's values depend only on the information present in certain local patches of the image. This is known as the \emph{field-of-view} problem \cite{fov}. The value of a neuron in a convolutional layer only depends on a certain subset of the input. This region in the input is the receptive field for that neuron. In learning tasks with images, each neuron in the output layer should ideally have a big receptive field, i.e. the values of the output neurons should indirectly depend on large parts of the image so that no important information of the image is left out when making the predictions. The receptive field size of a neuron can be increased, for example, by stacking more convolutional layers and thereby making the network deeper. This enlarges the receptive field size linearly in theory. Meanwhile, pooling or subsampling allows each pixel to cover a larger area and facilitates information travel over larger distances, which increases the receptive field size multiplicatively. However, just stacking up convolutional and pooling layers can have several disadvantages. In a deeper network, for example, the optimization problem \eqref{eq:invprob} might be hard to solve and repeatedly performing local operations can be computationally inefficient. 

In this article, we propose to employ additional layers which involve {\em nonlocal} interactions. Long-range and nonlocal dependencies are ubiquitous in several fields, such as financial markets \cite{long1}, physical sciences \cite{long3}, image denoising \cite{deblur} or image regularization \cite{deblur6}. We thus look at a few integral-based spatially nonlocal operators that could be used in CNNs. The underlying and associated partial integro-differential equation (PIDE) is then discretized in a so-called nonlocal block and inserted in the neural network. By direct computation of interactions between each pair of pixels or pair of patches in an image, these operators make sure that the field-of-view increases drastically. Consequently, the information travels larger distances for a fixed number of layers in a neural network.
\subsection{The nonlocal diffusion operator}
\begin{definition}
Let $\Omega \subseteq \mathbb{R}^n$ and let $ v:\mathbb{R}^n \rightarrow \mathbb R^n$ be any function, then for $\mathbf x, \mathbf y\in \Omega$, the nonlocal diffusion operator can be defined as
\begin{equation}\label{eq:nlop}
\mathcal L v(\mathbf x) := \int\limits_{\Omega} \omega(\mathbf x, \mathbf y) [ v(\mathbf y)- v(\mathbf x)] \,d\mathbf y,
\end{equation} where the kernel $\omega$ has the following two properties: (i) $\omega$ is symmetric: $\omega(\mathbf x, \mathbf y) = \omega(\mathbf y, \mathbf x), \\ \forall \mathbf x,\mathbf y \in \Omega$ and (ii) $\omega$ is positivity-preserving: $\omega(\mathbf x, \mathbf y) \geq 0, \forall \mathbf x,\mathbf y \in \Omega$.
\end{definition} The kernel can be viewed as a function that measures the affinity or similarity between points in $\Omega$. The aim is to apply such a spatial-only operator on a discrete image $v$, where $v:\Omega \rightarrow \mathbb{R}^C$ is the discrete image with $C$ channels, and $v(\mathbf x)$ is the set of pixel values of all the channels at a discrete spatial point $\mathbf x\in \Omega$ from the image domain $\Omega$ (the pixel strips in Figure \ref{fig:tstrip}).

Such operators have been an upshot of several image processing tasks. For example, for $\mathbf x, \mathbf y \in \Omega$, consider the functional $J( u, t) := \frac{1}{4} \int\limits_{\Omega \times \Omega} \omega(\mathbf x, \mathbf y) [ u(\mathbf x, t) -  u(\mathbf y,t)]^2 \, d\mathbf x \, d\mathbf y$ which sums up weighted variations of the function $u$, which is similar to total variational denoising in image processing. While minimizing the functional, the corresponding Euler-Lagrange descent flow is given by\begin{equation}
u_t(\mathbf x, t) = -J^\prime (u)(\mathbf x, t) =  -\int\limits_{\Omega} \omega(\mathbf x, \mathbf y)[u(\mathbf x, t)-u(\mathbf y, t)] \, d\mathbf y = \mathcal Lu(\mathbf x, t).
\end{equation}Therefore, starting from an initial input image $u_0(\mathbf x)$, the associated partial integro-differential equation (PIDE) can be written as\begin{equation}\label{eq:nlpide}
u_t(\mathbf x, t) = \mathcal Lu(\mathbf x, t), \quad u(\mathbf x, 0) = u_0(\mathbf x),
\end{equation} which is essentially a nonlocal weighted linear diffusion equation. The kernel function $\omega:\Omega \times \Omega \rightarrow \mathbb R$ is here assumed to be of Hilbert-Schmidt type \cite{hilbertschmidt}, i.e. $\int\limits_{\Omega} \int\limits_{\Omega} \vert \omega (\mathbf x, \mathbf y)\vert^2 \, d\mathbf x \, d\mathbf y < \infty$. There are several interpretations of the nonlocal diffusion operator that arise, e.g., from a continuous generalization of graph Laplacians or from a random walk Markov chain on $\Omega$.

The nonlocal diffusion operator $\mathcal L$ has many properties that are similar to a typical elliptic operator, such as $-\mathcal L$ being positive semi-definite. It can be shown that the system governed by equation \eqref{eq:nlpide} has energy decay \cite{deblur6}, i.e. \[\displaystyle\frac{d}{dt} \Vert u(\mathbf x, t)\Vert_{L^2}^2 = \frac{d}{dt}\int\limits_\Omega u(\mathbf x, t)^2\,d\mathbf x  \leq 0 \quad \text{ for all }\mathbf x \in \Omega, \quad \frac{d}{dt} \var[u(\mathbf x, t)]\leq 0,\] where $u(\mathbf x, t)$ is the solution of equation \eqref{eq:nlpide}. Since the operator $\mathcal L$ performs diffusion, the features tend to smooth out over time, and hence reduce the variance of the image/features. We will see later that, though the nonlocal interactions help in image classification, over-using it can damp the features leading to loss of information (see Section \ref{sec:msnb}).

Next, we look at a few pseudo-differential operators that can also be represented as integral-based global operators whose implementations introduce nonlocality into the neural network. The aim is to define a spatial operator $\mathcal O$ such that the associated forward propagation for a given input image (or set of feature maps) $u_0(\mathbf x)$ is governed by a PIDE that is similar to the form \begin{equation} \label{eq:nlpide2}
u_t(\mathbf x, t) = \mathcal Ou(\mathbf x, t), \quad u(\mathbf x, 0) = u_0(\mathbf x).
\end{equation}For instance, for a propagation following \eqref{eq:nlpide}, we have ${\mathcal O} = {\mathcal L}$.
\subsection{The fractional Laplacian operator} \label{sec:fraclap}
Fractional differential operators appear in several interesting problems in physical sciences \cite{fdiff1} or in financial mathematics \cite{fdiff5}. Let us first define a few related concepts before we define the operator itself.

\begin{definition}
Let $\Omega \subseteq \mathbb R^n$, $0<s<1$. The fractional-order Sobolev space is defined as \[ H^s(\Omega) :=  \bigg\{v \in L^2(\Omega) : \int\limits_\Omega \int\limits_\Omega \frac{[v(\mathbf y)-v(\mathbf x)]^2}{\vert \mathbf x-\mathbf y\vert^{n+2s}} \, d\mathbf y \, d\mathbf x < \infty \bigg\},\]
where the Gagliardo seminorm and the norm are defined as \[\vert v \vert_{H^s(\Omega)}^2 := \int\limits_\Omega \int\limits_\Omega \frac{[v(\mathbf y)-v(\mathbf x)]^2}{\vert \mathbf x-\mathbf y\vert^{n+2s}} \, d\mathbf y \, d\mathbf x, \quad \Vert v \Vert_{H^s(\Omega)} := \Big(\Vert v \Vert_{L^2(\Omega)}^2 + \vert v \vert_{H^s(\Omega)}^2\Big)^{1/2}.\]
\end{definition} 

\begin{definition} \label{def:fraclap}
For $v\in \mathcal S(\mathbb R^n)$, i.e. the Schwartz space of rapidly decreasing $C^\infty$ functions on $\mathbb R^n$, $0<s<1$, the fractional Laplacian operator $(-\Delta)^s$ is defined as the singular integral operator \begin{equation} \label{eq:fraclap}
(-\Delta)^s v(\mathbf x) := c_{n,s} \, P.V. \displaystyle \int\limits_{\mathbb R^n} \frac{[v(\mathbf x) - v(\mathbf y)]}{\vert \mathbf x - \mathbf y\vert^{n+2s}}  \, d\mathbf y = c_{n,s} \, \lim_{\epsilon\to 0^+} \displaystyle \int\limits_{\mathbb R^n\setminus B_\epsilon(\mathbf x)} \frac{[v(\mathbf x) - v(\mathbf y)]}{\vert \mathbf x - \mathbf y\vert^{n+2s}}  \, d\mathbf y, 
\end{equation}where P.V. stands for the Cauchy principal value (of the improper integral), $c_{n,s}$ is a constant depending on $n$, $s$ and is defined as $c_{n,s} = \frac{4^s \Gamma(n/2 +s)}{\pi^{n/2}\vert\Gamma(-s)\vert}$. 
\end{definition} 

Note that this operator can be seen as a special case of the nonlocal diffusion operator defined by equation \eqref{eq:nlop}, with kernel $\omega(\mathbf x, \mathbf y) = \vert\mathbf x - \mathbf y\vert^{-(n+2s)}$ and with the flipped term $v(\mathbf x) - v(\mathbf y)$ instead of $v(\mathbf y) - v(\mathbf x)$. The term is flipped because we want both operators to be elliptic, and the corresponding Poisson problem for the nonlocal diffusion operator is written as $-\mathcal L v(\mathbf x) = 0$, i.e. with an extra negative sign in front, whereas for the pseudo-differential operator, we have $(-\Delta)^s v(\mathbf x)= 0$.

\subsubsection*{Connection to pseudo-differential operators}
The operator \eqref{eq:fraclap} can in fact be defined as a \emph{pseudo-differential} operator (see Section SM1 in the supplementary material for further details), which shows the origins of the operator and also characterizes the operator in the Fourier space. Let the Fourier transform and the inverse Fourier transform of $v\in \mathcal S(\mathbb R^n)$ be denoted by $\hat v(\xi):= \mathscr F v(\xi)$ and $v(\mathbf x) := \mathscr F^{-1} \hat v(\mathbf x)$, respectively. It can be then shown that the fractional Laplacian is the pseudo-differential operator \cite{hitch} with the Fourier symbol $(2\pi\vert\xi\vert)^{2s}$ and satisfies \begin{equation}\label{eq:foulap}
\mathscr F((-\Delta)^s v)(\xi) = (2\pi\vert\xi\vert)^{2s} \, \hat v(\xi) \quad \text{ for } 0<s\leq 1.
\end{equation}For $s=1$, we get the classical Laplacian $(-\Delta)$ with the Fourier symbol $(2\pi \vert\xi\vert)^2$, i.e.\begin{align*}
-\Delta v(\mathbf x) &= -\Delta (\mathscr F^{-1}\hat v(\mathbf x)) =  -\Delta\Big[ \int\limits_{\mathbb R^n}\hat v(\xi) e^{2\pi i\mathbf x\cdot \xi}\, d\xi \Big]\\
&= \int\limits_{\mathbb R^n} (2\pi\vert\xi\vert)^2\,\hat v(\xi) e^{2\pi i \mathbf x\cdot \xi} \, d\xi = \mathscr F^{-1}\bigg((2\pi\vert\xi\vert)^2 \, \hat v(\xi)\bigg).
\end{align*}For classical PDE operators, the corresponding Fourier symbol is generally a polynomial in $\xi$, but by using symbols that are not only a function of $\xi$ but also a function $\mathbf x$, we arrive at a large plethora of pseudo-differential operators. It can be shown that the classical Laplacian is a limit case of the fractional one \cite{stinga}, i.e. for a certain class of bounded functions $v$, $\mathbf x\in \mathbb R^n$, we have the point-wise limits $\lim_{s\to 0^+}(-\Delta)^s v(\mathbf x)=  v(\mathbf x)$, $\lim_{s\to 1^-} (-\Delta)^s v(\mathbf x) = -\Delta v(\mathbf x)$. Note that there are several other equivalent definitions of the fractional Laplacian operator that are defined via heat semi-groups \cite{fracdef1}, or as an inverse of the Riesz potential, etc. (see \cite{fdiff3}). For our purposes of image classification and segmentation, we will be using the singular integral operator from Definition \ref{def:fraclap} that defines the output of the operator point-wise.

\subsection{The inverse fractional Laplacian operator}
Now we define an analogous operator $(-\Delta)^{-s}$ for $s>0$, i.e. an inverse fractional Laplacian, which, as we will see, is also a nonlocal operator. Note that, similar to equation \eqref{eq:foulap}, we have \begin{equation}\label{eq:fractop}\mathscr F((-\Delta)^{-s} v)(\xi) = (2\pi\vert\xi\vert)^{-2s} \, \hat v(\xi) \quad \text{ for } 0<s< n/2.\end{equation}
One needs the restriction $0<s<n/2$ because if $s\geq n/2$, the Fourier symbol/multiplier $(2\pi\vert\xi\vert)^{-2s}$ fails to define a \emph{tempered distribution} \cite{fdiff5}. The symbol $(2\pi\vert\xi\vert)^{-2s}$ is decaying with respect to $s$, and, in the sense of distributions (see \cite{riesz3}), such symbols have the Fourier inverse $\mathscr F^{-1} (\vert\xi\vert^{-2s}) = c_{n,-s}\vert \mathbf x \vert^{-(n-2s)}$. Using the inverse property for convolutions $f\ast g = \mathscr F^{-1} \{\mathscr F\{f\} \cdot \mathscr F\{g\} \}$ and equation \eqref{eq:fractop}, we get the following definition.
\begin{definition} \label{def:invfraclap}
For $v\in \mathcal S(\mathbb R^n)$, $0<s<n/2$, the inverse fractional Laplacian operator $(-\Delta)^{-s}$ is defined as the integral operator \begin{equation} \label{eq:invfraclap}
(-\Delta)^{-s} v(\mathbf x) := \Big( c_{n,-s} \vert \mathbf x\vert^{-(n-2s)}\Big) \ast v(\mathbf x) = c_{n,-s} \int\limits_{\mathbb R^n} \frac{v(\mathbf y)}{\vert \mathbf x - \mathbf y\vert^{n-2s}} \, d\mathbf y,
\end{equation}where $c_{n,-s}$ is the constant given by $c_{n,-s} = \frac{ \Gamma(n/2 -s)}{4^s \pi^{n/2}\Gamma(s)}$.
\end{definition}The right-hand side of the definition is called the \emph{Riesz potential}. This singular integral is well-defined, provided $v$ decays sufficiently rapidly at infinity (see \cite{riesz3}), for instance, if $v \in L^p(\mathbb R^n)$ with $1 \leq p < \frac {n}{2s}$.

Recently, a fundamental solution for the case $s=n/2$ has been proposed \cite{stinga2}, which can also be used in the neural network as a nonlocal operation. As pointed out in \cite{riesz3}, the logarithmic kernel defined below can be seen as a vague limit of the Riesz potentials defined by \eqref{eq:invfraclap}.
\begin{definition}
Let $v\in \mathcal S(\mathbb R^n)$ with $\int\limits_{\mathbb R^n} v = 0$, $\mathbf x\in \mathbb R^n$, and $s=n/2$. Then we have
\begin{equation}\label{eq:invfraclap2}
(-\Delta)^{-s}v(\mathbf x) := c_n \int\limits_{\mathbb R^n}\bigg(-2 \log \vert \mathbf x - \mathbf y\vert - \gamma\bigg) v(\mathbf y)\, d\mathbf y,
\end{equation}where the Euler-Mascheroni constant $\gamma$ is given by $\gamma=-\int\limits_0^\infty e^{-r} \log r \, dr \approx 0.5772156649$, and the constant $c_n$ is given by $c_n =  \frac{1}{(4\pi)^{n/2} \Gamma(n/2)}$.
\end{definition}

As such, the Riesz potential kernel $\frac{1}{\vert \mathbf x - \mathbf y\vert^{n-2s}}$ may not always satisfy the square-integrability condition. Besides, in the case of the $\log$ kernel from equation \eqref{eq:invfraclap2}, the kernel is no longer positivity-preserving. While implementing these nonlocal operators for images, methods used to bypass this issue and to avoid the blowing up of values are discussed in Sections \ref{sec:imfraclap1} and \ref{sec:imfraclap2}. It is interesting to note that the kernels of the pseudo-differential operators are translation invariant and isotropic. Also, from Definition \ref{def:invfraclap}, it is clear that the operators from equations \eqref{eq:invfraclap} and \eqref{eq:invfraclap2} can be seen as convolutions on a global scale (spatially), whereas the operators $\mathcal L$ and $(-\Delta)^s$ from equations \eqref{eq:nlop} and \eqref{eq:fraclap}, respectively, define a diffusion-like operator on a global scale (spatially).

\section{Implementation details}
In this section, we first look at the implementation of the Hamiltonian network that will be used as the base network. Then, several discretization details about the nonlocal blocks, which contain the proposed integral operators, will be introduced. We explore a few strategies to reduce the computational effort of such global operators and discuss how the respective tensors are dealt with when such operators are applied on images or feature maps. Finally, the general implementation details for the numerical experiments, including the regularization techniques, are discussed.
\subsection{Hamiltonian networks with Verlet scheme}
The Hamiltonian network will be used as the base architecture to ensure stability of the forward propagation, and changes are made to it by introducing nonlocality in the network. Similar to the ResNet architecture, several of the Hamiltonian blocks \cite{rev} are stacked up to form a Unit and several of these reversible Units together form the entire network. Figure \ref{fig:hamnet} shows a 2-Unit Hamiltonian network as an example, where we see that the input is passed through an initial convolution layer and then through each Unit of the network. 
\input{images/hamnet}
Similar to the original ResNet architecture \cite{he2016}, after each Unit (of $m$ Hamiltonian blocks), a pooling operation reduces the feature map size by half, and the number of channels is increased by using a $1\times 1$ convolution layer (with ReLU). After the final Unit, the features are subsampled and passed on to the fully-connected layer (for image classification tasks), where the predictions are made. The number of Units is kept fixed. 

In our case, the networks will always contain three Units, each of which has $m$ Hamiltonian blocks. In terms of the ODE/PDE interpretation, each block represents a time step of the discretization, and the final output $\mathbf Y_N$ is the output of the last block in this network. Therefore, a normal Hamiltonian network has $12m+2$ layers, with $4m$ layers in each Unit (2 convolutions and 2 transposed convolutions per Hamiltonian block).
To include the global operators into the network the nonlocality is added after the second Hamiltonian block in each Unit, i.e.\ the feature maps obtained after the second block are passed through the \emph{nonlocal block}. The output of the nonlocal block is then fed into the next \emph{normal} Hamiltonian block. Hence, altogether we have three additional nonlocal blocks in our network, one in each Unit.
\subsection{Nonlocal diffusion operator}\label{sec:implement}
Let $\mathbf X \in \mathbb R^{H\times W\times C}$ be the input to the \emph{nonlocal block} that is supposed to contain the nonlocal interactions. Let $\mathbf X_i$ denote each pixel of the image, i.e. for an input $\mathbf X$ with $C$ channels, $\mathbf X_i\in \mathbb R^{C}$ stands for the $C$ channel values for each pixel position. These vectors will be called \emph{pixel strips} from here on. To implement the nonlocal diffusion operator $\mathcal L$ defined by equation \eqref{eq:nlop} in the nonlocal block, we discretize the PIDE-inspired operator using a 2-step computation, which strengthens the nonlocal interactions between the features: 

\begin{gather}\label{eq:nldes}
\begin{aligned}
\mathbf [\mathbf B_1]_i &= \mathbf X_i + h \, \mathcal K_1 \Bigg[\displaystyle\frac{1}{\displaystyle\sum_j \omega(\mathbf X_i, \mathbf X_j)}\,\mathlarger{\mathlarger{\sum}}_j \omega(\mathbf X_i, \mathbf X_j) (\mathbf X_j - \mathbf X_i)\Bigg],\\
[\mathbf B_2]_i &= \mathbf X_i + h \, \mathcal K_2 \Bigg[ \displaystyle\frac{1}{\displaystyle\sum_j \omega(\mathbf X_i, \mathbf X_j)}\, \mathlarger{\mathlarger{\sum}}_j \omega(\mathbf X_i, \mathbf X_j) \Big([\mathbf B_1]_j - [\mathbf B_1]_i\Big) \Bigg],
\end{aligned}\end{gather}where $[\mathbf B_1]_i$ and $[\mathbf B_2]_i$ are the individual pixel strips of the intermediate feature maps $\mathbf B_1$ and the output feature maps $\mathbf B_2$ respectively, $h$ is the discretization step size, and $\mathcal K_1$, $\mathcal K_2$ are $1\times 1$ convolution operators, followed by ReLU activation and batch normalization. The output $\mathbf B_2$ is then fed into the next normal block in the Unit.

\subsubsection*{Relation to Verlet time-discretization} 
Let us investigate how \eqref{eq:nldes} can be interpreted as a Verlet time-discretization of a system of ODEs, which represent a spatial discretization of the PIDE \eqref{eq:nlpide2}.
To this end, set $\mathbf V^{\text{old}} := \mathbf X^{\text{old}} := \mathbf X \in \mathbb{R}^{C \times H W}$ and $\mathbf W^{\text{old}} := - \mathbf X^{\text{old}} \in \mathbb{R}^{C \times HW}$. Now, for $i=1,\ldots,H \times W$, we define a two-step iteration scheme by
\begin{gather}\label{eq:nldes_revisited}
\begin{aligned}
\mathbf V_i^{\text{new}} &= \mathbf V_i^{\text{old}} + h \, \mathcal K_1 \Bigg[\displaystyle\frac{1}{\displaystyle\sum_j \omega(\mathbf V_i^{\text{old}}, \mathbf V_j^{\text{old}})}\,\mathlarger{\mathlarger{\sum}}_j \omega(\mathbf V_i^{\text{old}}, \mathbf V_j^{\text{old}}) (\mathbf W_i^{\text{old}} - \mathbf W_j^{\text{old}})\Bigg],\\
\mathbf W_i^{\text{new}} &=  \mathbf W_i^{\text{old}} - h \, \mathcal K_2 \Bigg[ \displaystyle\frac{1}{\displaystyle\sum_j \omega(-\mathbf W_i^{\text{old}}, -\mathbf W_j^{\text{old}})}\, \mathlarger{\mathlarger{\sum}}_j \omega(-\mathbf W_i^{\text{old}}, - \mathbf W_j^{\text{old}}) \Big(\mathbf V_j^{\text{new}} - \mathbf V_i^{\text{new}}\Big) \Bigg].
\end{aligned}
\end{gather}We directly see that $\mathbf V_i^{\text{new}} = [\mathbf B_1]_i$ and 
$\mathbf X^{\text{new}}_i := -\mathbf W_i^{\text{new}} = [\mathbf B_2]_i$. Moreover, \eqref{eq:nldes_revisited} is a Verlet scheme for a specific system of PIDEs as we will see now. To this end, let the $\mathbb{R}^{HW \times HW}$-valued operators $\Omega_1, \Omega_2$ be defined by
\begin{equation*}
 \Big[\Omega_1(\mathbf X) \Big]_{jk} := \frac{\omega(\mathbf X_j, \mathbf X_k)}{\sum_{m} \omega( \mathbf X_k, \mathbf X_m)} ~~~ \text{ and } ~~~  \Big[\Omega_2(\mathbf X) \Big]_{jk} := \frac{\omega(- \mathbf X_j, - \mathbf X_k)}{\sum_{m} \omega(- \mathbf X_k, - \mathbf X_m)}.
\end{equation*}
Note that $\Omega_1(\mathbf V^{\text{old}}) = \Omega_2(\mathbf W^{\text{old}})$. Furthermore, we define the operator $\operatorname{Diff}: \mathbb{R}^{C \times HW} \to \mathbb{R}^{HW \times HW \times C}$ by $\Big[\operatorname{Diff}(\mathbf X)\Big]_{jk} = \mathbf X_k - \mathbf X_j$. With these operators we rewrite \eqref{eq:nldes_revisited} as matrix/tensor equation
\begin{gather*}
\begin{aligned}
\mathbf V^{\text{new}} &= \mathbf V^{\text{old}} + h \, \mathcal K_1 \Bigg[\operatorname{diag}_{12}\Big(\Omega_1(\mathbf V^{\text{old}}) \operatorname{Diff}(\mathbf{W}^{\text{old}}) \Big)^T\Bigg],\\
\mathbf W^{\text{new}} &=  \mathbf W^{\text{old}} - h \, \mathcal K_2 \Bigg[ - \operatorname{diag}_{12}\Big(\Omega_2(\mathbf W^{\text{old}}) \operatorname{Diff}(\mathbf{V}^{\text{new}})\Big)^T \Bigg],
\end{aligned}\end{gather*}
where $\operatorname{diag}_{12}: \mathbb{R}^{HW \times HW \times C} \to \mathbb{R}^{HW \times C}$ denotes the diagonal operator with respect to the first two dimensions, i.e.\ 
$[\operatorname{diag}_{12}(\mathbf Y)]_{jk} := \mathbf Y_{jjk}$. Note that the multiplication of $\Omega_{1}$ or $\Omega_2$ with the output of the $\operatorname{Diff}$ operator has to be understood in analogy to a multiplication between a $HW \times HW$ matrix and a $HW \times HWC$ matrix. With this, we can finally define the nonlinear system of ODEs
\begin{gather}\label{eq:PIDEsystem}
\begin{aligned}
\dot{\mathbf V}(t) &=  \mathcal{K}_1(t)\Bigg(\operatorname{diag}_{12}\Big(\Omega_1(\mathbf V(t)) \operatorname{Diff}(\mathbf W(t))\Big)^T\Bigg),\\
\dot{\mathbf W}(t) &= -\mathcal{K}_2(t)\Bigg(-\operatorname{diag}_{12} \Big(\Omega_2(\mathbf W(t)) \operatorname{Diff}(\mathbf V(t))\Big)^T\Bigg).
\end{aligned}
\end{gather}
Now it is easy to observe that the Verlet time-discretization of this system results in \eqref{eq:nldes_revisited}. The latter resembles a spatially discretized variant of 
\begin{equation*}
u_t(\mathbf x, t) = \mathcal Ou(\mathbf x, t), \quad u(\mathbf x, 0) = u_0(\mathbf x),
\end{equation*}
with ${\mathcal O} = {\mathcal L}$ and an additional outer convolution by $\mathcal{K}_1$ and $\mathcal{K}_2$. This has to be seen as an analogy to the system \eqref{eq:ham} and its time-discretization \eqref{eq:verlet} from \cite{rev}. Thus, our nonlocal block iteration \eqref{eq:nldes} resembles the Verlet discretization of a specific spatially discretized PIDE.


\ifx
\begin{equation} \label{eq:pid}
u_t(\mathbf x, t) = \mathcal O(u(\mathbf x) + h\mathcal L u(\mathbf x)), \quad u(\mathbf x, 0) = u_0(\mathbf x),
\end{equation} where $u_0(\mathbf x)$ is the initial value and $\mathcal O$ is one of the nonlocal operators defined before.
\fi

\subsubsection*{Affinity kernel and outer convolution}
\begin{figure}[h!]
  \centering
  \begin{minipage}[b]{0.49\textwidth}
    \includegraphics[width=\textwidth]{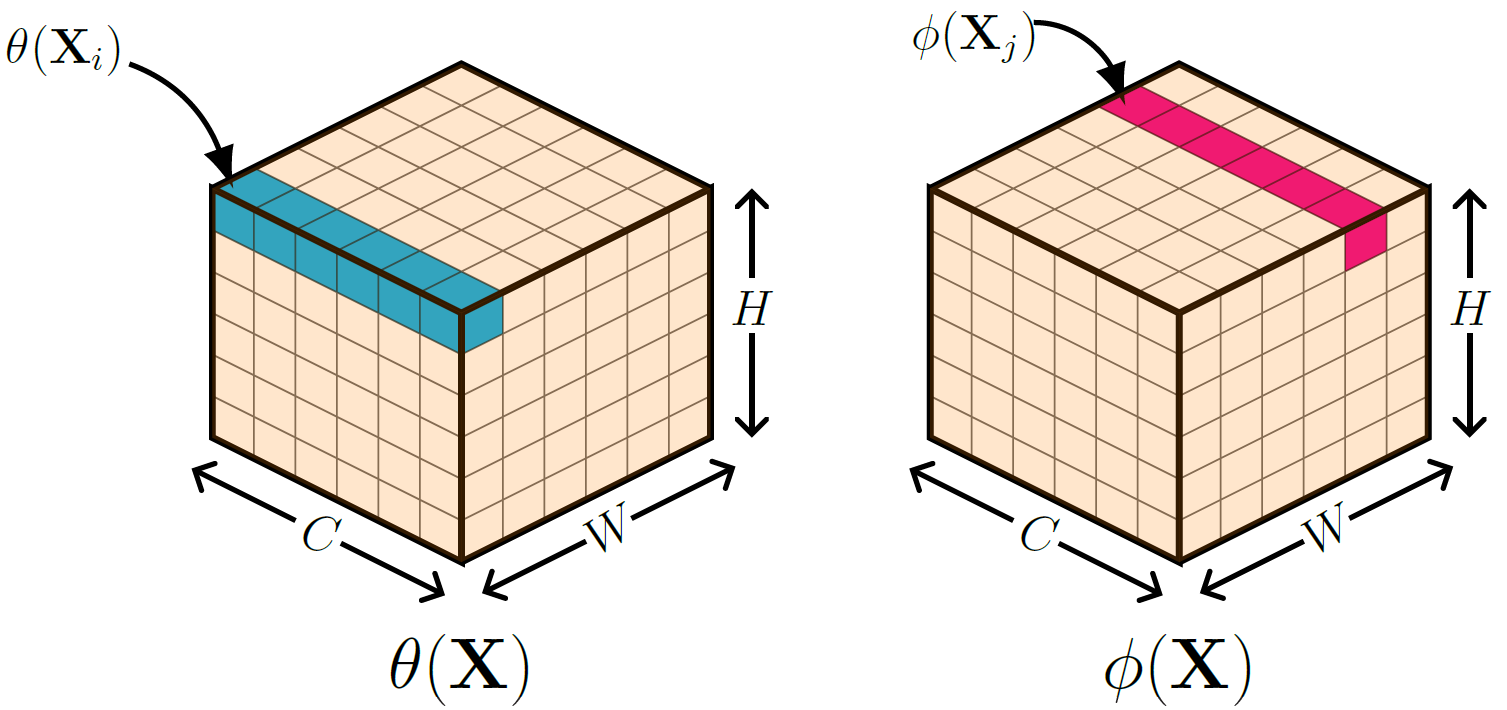}
  \end{minipage}
  \hfill\vline\hfill
  \begin{minipage}[b]{0.49\textwidth}
    \includegraphics[width=\textwidth]{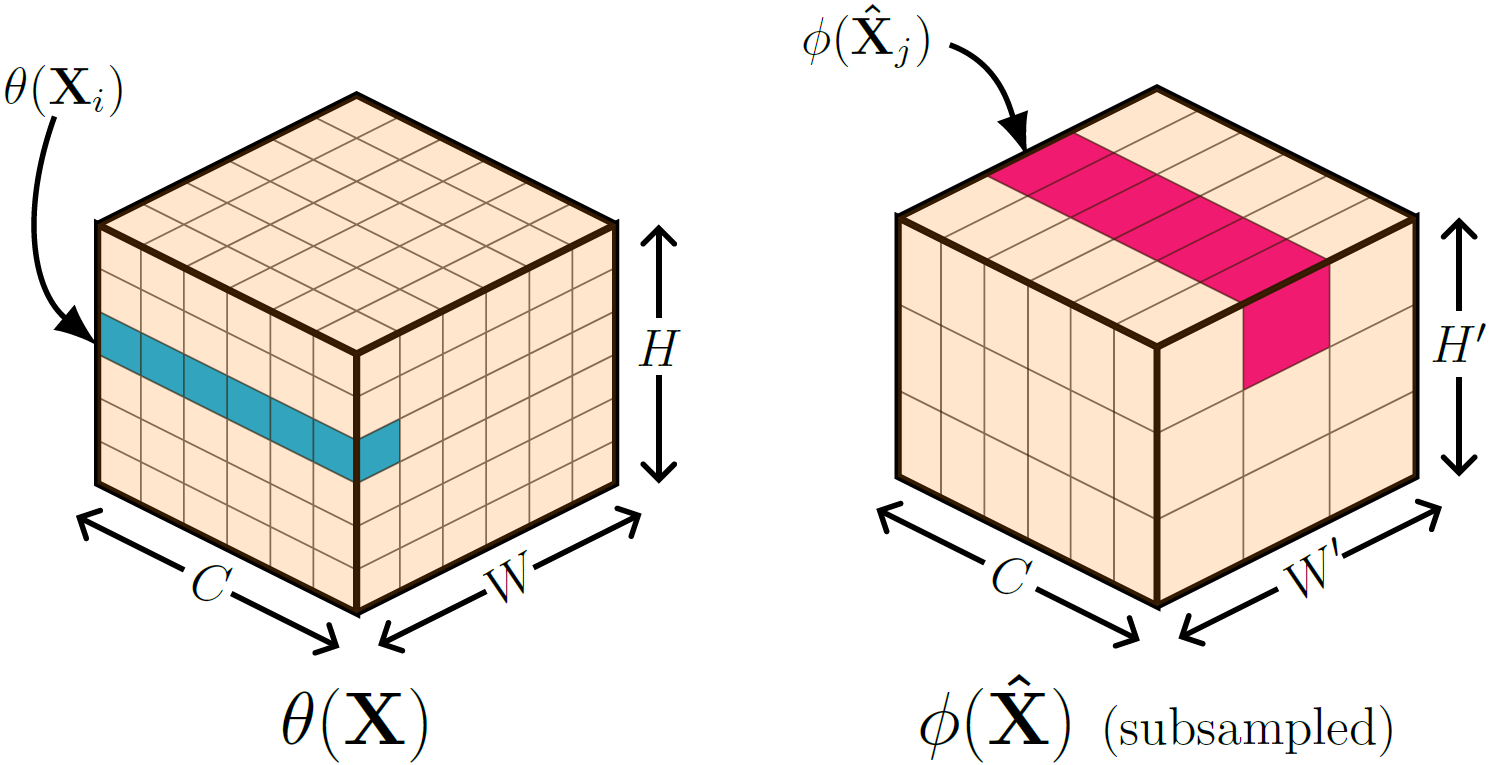}
  \end{minipage}
 \caption{Pixel strips of embedded versions of $\mathbf X$ with $C$ channels, which are used to compute the unsubsampled kernel entry $\omega(\mathbf X_i, \mathbf X_j)$ (\textit{left}) and the subsampled version of the kernel entry $\omega(\mathbf X_i, \mathbf{\hat X}_j)$ (\textit{right}).} \label{fig:tstrip} 
\end{figure}

There are several things to note here regarding the kernel and the $1\times 1$ convolution. The kernel $\omega(\mathbf x, \mathbf y)$ measures the affinity or similarity between each pair of pixel strips of the image/feature maps. For choosing the right kernel, there are several suggestions made by Wang, Girshick, Gupta et al. in \cite{helocal}, such as the Gaussian function $e^{\mathbf x^T \mathbf y}$, the embedded Gaussian $e^{\theta(\mathbf x)^T \phi(\mathbf y)}$ and the embedded dot product $\theta(\mathbf x)^T \phi(\mathbf y)$. All these kernels seem to work equally well. We will be using the scaled embedded dot product $\lambda\theta(\mathbf x)^T \phi(\mathbf y)$, where $\theta$ and $\phi$ are $1\times 1$ convolutional embeddings, and $\lambda>0$ is a scaling factor. Essentially, it means that the input $\mathbf X$ is passed through two different $1\times 1$ convolutions to obtain embeddings $\theta(\mathbf X)$ and $\phi(\mathbf X)$. Then the kernel $\omega$ measures the affinity between the pairs of pixel strips $\theta(\mathbf X_i)$ and $\phi(\mathbf X_j)$ (see Figure \ref{fig:tstrip}). To reduce complexity, subsampling, e.g. pooling, can be applied within a pixel strip, see Figure \ref{fig:tstrip} (right).

One can also use the scaled Gaussian kernel if one needs a positivity-preserving kernel $e^{\lambda\theta(\mathbf x)^T \phi(\mathbf y)}$. Note that the embedded kernels are not symmetric. If one needs a symmetric kernel with embeddings of the input, then one can pre-embed the features to get $\theta(\mathbf X)$ and then feed it to the PIDE-inspired nonlocal block with the kernel $\omega(\mathbf X_i, \mathbf X_j)$ in \eqref{eq:nldes} replaced by $\omega(\theta(\mathbf X_i), \theta(\mathbf X_j))$. Then one can use the scaled Gaussian kernel $\omega(\mathbf x, \mathbf y)= e^{\lambda\mathbf x^T \mathbf y}$ or the scaled dot product $\omega(\mathbf x, \mathbf y)=\lambda \mathbf x^T \mathbf y$, both of which are symmetric.
The term $\textstyle \sum_j \omega(\mathbf X_i, \mathbf X_j)$ is a normalizing factor. For the sake of simplicity, and for easier gradient computations, as suggested in \cite{helocal}, we choose this factor to be $\mathscr N$, where $\mathscr N=H\times W$ is the number of pixel strips in the input $\mathbf X$. For instance, in Figure \ref{fig:tstrip}, the number of pixel strips is $36$ for a $6\times 6$ image. Without this normalizing factor, the computations might blow up for inputs of larger sizes.

\begin{figure}[h!]
	\centering
\tdplotsetmaincoords{60}{30}
\resizebox{.9\linewidth}{!}{\begin{tikzpicture}[tdplot_main_coords,line join=miter,font=\sffamily, scale = 0.45]
\begin{scope}[canvas is yz plane at x=-0.8+0.4*1]
   \pgfmathtruncatemacro{\fullness}{120-20*1}
   \draw[fill=orange!\fullness] (-1,-1) rectangle (1,1);
  \end{scope}
  \begin{scope}[canvas is yz plane at x=-0.8+0.4*2]
   \pgfmathtruncatemacro{\fullness}{120-20*2}
   \draw[fill=orange!\fullness] (-1,-1) rectangle (1,1);
  \end{scope}
    \begin{scope}[canvas is yz plane at x=-0.8+0.4*3]
   \pgfmathtruncatemacro{\fullness}{120-20*3}
   \draw[fill=orange!\fullness] (-1,-1) rectangle (1,1);
  \end{scope}
      \begin{scope}[canvas is yz plane at x=-0.8+0.4*4]
   \pgfmathtruncatemacro{\fullness}{120-20*4}
   \draw[fill=orange!\fullness] (-1,-1) rectangle (1,1);
  \end{scope}
        \begin{scope}[canvas is yz plane at x=-0.8+0.4*5]
   \pgfmathtruncatemacro{\fullness}{120-20*5}
   \draw[fill=orange!\fullness] (-1,-1) rectangle (1,1);
  \end{scope}
  \node[anchor=center] at (-1,-1) {\fontsize{3pt}{3.6pt}\selectfont ${\mathbf X}$}; 
  
  \tdplotsetrotatedcoords{-60}{0}{0}
  \draw[black, -{Latex[scale=0.7]}, tdplot_rotated_coords, anchor = center] (0,1.25,0) -- (0,3,0);
  	\node[black,  tdplot_rotated_coords, anchor = center] at (0, 2.15, 0.34) {\fontsize{3pt}{3.6pt}\selectfont ${\mathbf X}$};
  \draw[thin, draw=black, fill=red!50!blue, fill opacity=0.3, tdplot_rotated_coords] (0,3,1.5) rectangle (0, 3.8,-1.5);
  \node[black, rotate = 90,  anchor=center, tdplot_rotated_coords] at (0, 3.4, 0) {\tiny Stage $1$};
    \draw[->, -{Latex[scale=0.7]}, black, tdplot_rotated_coords] (0,3.8,0) -- (0,5.25,0);
    \node[black,  tdplot_rotated_coords, anchor = center] at (0, 4.45, 0.34) {\fontsize{3pt}{3.6pt}\selectfont ${h}$};
   \draw[black, fill=green!50!yellow, fill opacity=0.4, tdplot_rotated_coords] (0, 5.65,0) circle (12pt);
   \node [anchor = center, black, tdplot_rotated_coords] at (0,5.65,0) {\LARGE $+$};   
   \draw[->, -{Latex[scale=0.7]}, black, tdplot_rotated_coords] (0, 1.25, 0) .. controls (0, 2.4, -2.5) and (0,4.3,-2.1) .. (0, 5.7,-0.45);
   \draw[->, -{Latex[scale=0.7]}, black, tdplot_rotated_coords] (0,6.05,0) -- (0,7.5,0);
   \node[black,  tdplot_rotated_coords, anchor = center] at (0, 6.6, 0.34) {\fontsize{3pt}{3.6pt}\selectfont ${\mathbf B_1}$};
     \draw[thin, draw=black, fill=red!50!blue, fill opacity=0.3, tdplot_rotated_coords] (0,7.5,1.5) rectangle (0, 8.3,-1.5);
  \node[black, rotate = 90, anchor=center,   tdplot_rotated_coords] at (0, 7.9, 0) {\tiny Stage $2$};
  \draw[->, -{Latex[scale=0.7]}, black, tdplot_rotated_coords] (0,8.3,0) -- (0,9.75,0);
    \node[black,  tdplot_rotated_coords, anchor = center] at (0, 8.95, 0.34) {\fontsize{3pt}{3.6pt}\selectfont ${h}$};
   \draw[black, fill=green!50!yellow, fill opacity=0.4, tdplot_rotated_coords] (0, 10.15,0) circle (12pt);
   \node [anchor = center, black, tdplot_rotated_coords] at (0,10.15,0) {\LARGE $+$}; 
   \draw[->, -{Latex[scale=0.7]}, black, tdplot_rotated_coords] (0, 1.25, 0) .. controls (0, 2.4, -3.5) and (0,8.9,-3.5) .. (0, 10.2,-0.45);
   \draw[->, -{Latex[scale=0.7]}, black, tdplot_rotated_coords] (0,10.55,0) -- (0,12,0);
	\node[black,  tdplot_rotated_coords, anchor = center] at (0, 11.1, 0.34) {\fontsize{3pt}{3.6pt}\selectfont ${\mathbf B_2}$};
\begin{scope}[canvas is yz plane at x=13.5+0.4*1]
   \pgfmathtruncatemacro{\fullness}{120-20*1}
   \draw[fill=blue!35!magenta!35!red] (0,2.5) rectangle (2,4.5);
  \end{scope}
  \begin{scope}[canvas is yz plane at x=13.5+0.4*2]
   \pgfmathtruncatemacro{\fullness}{120-20*2}
   \draw[fill=blue!45!magenta!45!red] (0,2.5) rectangle (2,4.5);
  \end{scope}
    \begin{scope}[canvas is yz plane at x=13.5+0.4*3]
   \pgfmathtruncatemacro{\fullness}{120-20*3}
   \draw[fill=blue!55!magenta!55!red] (0,2.5) rectangle (2,4.5);
  \end{scope}
      \begin{scope}[canvas is yz plane at x=13.5+0.4*4]
   \pgfmathtruncatemacro{\fullness}{120-20*4}
   \draw[fill=blue!65!magenta!65!red] (0,2.5) rectangle (2,4.5);
  \end{scope}
        \begin{scope}[canvas is yz plane at x=13.5+0.4*5]
   \pgfmathtruncatemacro{\fullness}{120-20*5}
   \draw[fill=blue!75!magenta!75!red] (0,2.5) rectangle (2,4.5);
  \end{scope}
   
   \node[anchor=center, tdplot_rotated_coords] at (0,15,0) {\fontsize{3pt}{3.6pt}\selectfont ${\mathbf B_2}$};   
  \end{tikzpicture}}
	\caption{The two-stage computation of a Nonlocal Block with skip connections.}
	\label{fig:twostage}
\end{figure}
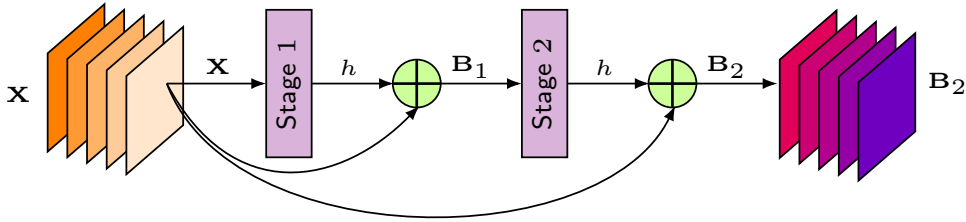

After the normalization, a $1\times 1$ convolution $\mathcal K_1$ is applied, along with ReLU and batch normalization, and then the result is added to the pixel strip of the original input, i.e. $\mathbf X_i$. Instead of simply discretizing the PIDE (equations \eqref{eq:nlpide} and \eqref{eq:nlpide2}) using the forward Euler scheme, this computation is iteratively repeated several times, as shown in Figure \ref{fig:twostage}. This allows us to extract more nonlocal information from the dataset.
The advantage of this multi-stage operation is that one can perform nonlocal operations several times, which widens the field-of-view of the network further, but the kernel $\omega$ is computed only once at the start, which saves computational time and effort. At the same time, one cannot have too many stages in one nonlocal block because, for instance, after Stage 5, we have the output $\mathbf {B}_5$, but the pre-computed kernel would fail to characterize the affinity between pairs of pixel strips of the features properly since the features would change quite a bit after each stage (see Section \ref{sec:msnb}). Here, we restrict ourselves to a two-stage forward propagation, i.e. we repeat the computation twice.

It is important to note that this nonlocal block differs from the one suggested in \cite{taolocal} because, as shown in Figure \ref{fig:twostage}, it contains identity skip connections. Similar to skip connections in ResNets, which have performed well for deep networks, skip connections in the nonlocal block are introduced to make sure that the nonlocal block does not impede the flow of information, which would force the network to perform worse. The experiments in Section \ref{sec:benchmark} confirm this, showing that including skip connections in the nonlocal computations can be beneficial. This way, in the presence of nonlocal blocks, the network at least does not deliver a worse performance than a network without the nonlocality in the worst case. The exact details of how the tensors are dealt with and propagated forward in the nonlocal block are shown in SM3 in the supplementary material.
\subsection{Fractional Laplacian operator $(-\Delta)^s$}\label{sec:imfraclap1}
As discussed before, the fractional Laplacian operator \eqref{eq:fraclap} can be seen as a special case of the nonlocal diffusion operator with kernel $\omega(\mathbf x, \mathbf y) = \vert\mathbf x - \mathbf y\vert^{-(n+2s)}$. Thus, the discretization can follow a similar procedure. In this case, we have the two-stage forward propagation\begin{gather}\label{eq:nldes2}
\begin{aligned}
\mathbf [\mathbf B_1]_i &= \mathbf X_i + h \, \mathcal K_1 \Bigg[\displaystyle\frac{c_{n,s}}{\displaystyle\sum_j \omega(\mathbf X_i, \mathbf X_j)}\,\mathlarger{\mathlarger{\sum}}_j \frac{\lambda}{\big\Vert\theta(\mathbf X_i) - \phi(\mathbf X_j)\big\Vert^{n+2s}_2} (\mathbf X_i - \mathbf X_j)\Bigg],\\
[\mathbf B_2]_i &= \mathbf X_i + h \, \mathcal K_2 \Bigg[ \displaystyle\frac{c_{n,s}}{\displaystyle\sum_j \omega(\mathbf X_i, \mathbf X_j)}\, \mathlarger{\mathlarger{\sum}}_j \frac{\lambda}{\big\Vert\theta(\mathbf X_i) - \phi(\mathbf X_j)\big\Vert^{n+2s}_2}  \Big([\mathbf B_1]_i - [\mathbf B_1]_j\Big) \Bigg],
\end{aligned}\end{gather}where $\theta$ and $\phi$ are again $1\times 1$ convolutional embeddings, $\Vert \cdot \Vert_2$ is the $L^2$ norm, $0<s<1$, and $\lambda>0$ is a scaling constant that is used to have greater control over the diffusion kernel. A small value of $\lambda$ leads to a weak interaction between the different pixel strips, whereas a large value of $\lambda$ damps the signal quickly. The rest of the expression is implemented in a similar fashion, with normalizing constant $\mathscr N$, etc (see Section SM3 in the supplementary material for further details). The forward propagation of the features through this 2-stage process can be again seen as a discretization of an underlying PIDE (equation \eqref{eq:nlpide2} with $\mathcal{O} = \mathcal{L}$).

Besides, we have a singularity in the kernel if the embeddings of the pixel strips, i.e. $\theta(\mathbf X_i)$ and $\phi(\mathbf X_j)$, are the same. From Section SM2 in the supplementary material, we know that there is no net contribution to the integral when $\mathbf y \to \mathbf x$. Hence we just perform a \emph{safe divide}, i.e. the kernel entry $\omega(\mathbf X_i, \mathbf X_j)=\frac{\lambda}{\big\Vert\theta(\mathbf X_i) - \phi(\mathbf X_j)\big\Vert^{n+2s}_2}$ is replaced with 0 if the denominator turns out to be zero.
\ifx
While computing the pairwise distances, if the mini-batch size is large, the tensors get replicated and thus consume a lot of memory. To overcome this problem, we can use the identity \[\Vert \mathbf x - \mathbf y \Vert^2=\Vert\mathbf x\Vert^2-2\mathbf x\cdot\mathbf y+\Vert\mathbf y\Vert^2.\] This reduces the memory footprint by quite a bit. For instance, if we have matrices $A\in \mathbb R^{36\times 6}$ and $B\in \mathbb R^{9\times 6}$ that contain the values of $36$ and $9$ pixel strips of the two embeddings respectively (see right side of Figure \ref{fig:tstrip}), then the square of the pairwise distance is roughly represented as follows

\[
\begin{pmatrix}
\vdots\\
\end{pmatrix}_{36\times 1}-2\big[A\cdot B^T\big] + \begin{pmatrix}
\hdots
\end{pmatrix}_{1\times 9},
\] with $36\times 1$ and $1\times 9$ representing the squared $L^2$ norms of the $36$ and $9$ pixel strips respectively. The addition and the subtraction is done row-wise and column-wise, not element-wise. For example, each of the 36 entries of the $36\times 1$ vector is added to the corresponding row of the $36\times 9$ matrix that results from the dot product operation.\fi

\subsection{Inverse fractional Laplacian operator $(-\Delta)^{-s}$} \label{sec:imfraclap2}
The inverse fractional Laplacian operator for $0<s<n/2$, defined by equation \eqref{eq:invfraclap}, can be discretized in a similar fashion using a two-stage method with skip connections, i.e.

\begin{gather}\label{eq:nldes3}
\begin{aligned}
\mathbf [\mathbf B_1]_i &= \mathbf X_i + h \, \mathcal K_1 \Bigg[\displaystyle\frac{c_{n,-s}}{\displaystyle\sum_j \omega(\mathbf X_i, \mathbf X_j)}\,\mathlarger{\mathlarger{\sum}}_j \frac{\lambda}{\big\Vert\theta(\mathbf X_i) - \phi(\mathbf X_j)\big\Vert^{n-2s}_2} \, \mathbf X_j\Bigg],\\
[\mathbf B_2]_i &= \mathbf X_i + h \, \mathcal K_2 \Bigg[ \displaystyle\frac{c_{n,-s}}{\displaystyle\sum_j \omega(\mathbf X_i, \mathbf X_j)}\, \mathlarger{\mathlarger{\sum}}_j \frac{\lambda}{\big\Vert\theta(\mathbf X_i) - \phi(\mathbf X_j)\big\Vert^{n-2s}_2}  \, [\mathbf B_1]_j \Bigg].
\end{aligned}\end{gather}For $s=n/2$, we have a $\log$ kernel \eqref{eq:invfraclap2}, which is implemented in the nonlocal block as
\begin{gather}\label{eq:nldes4}
\begin{aligned}
\mathbf [\mathbf B_1]_i &= \mathbf X_i + h \, \mathcal K_1 \Bigg[\displaystyle\frac{c_{n}}{\displaystyle\sum_j \omega(\mathbf X_i, \mathbf X_j)}\,\mathlarger{\mathlarger{\sum}}_j \bigg(-2\lambda\log\bigg(\big\Vert\theta(\mathbf X_i) - \phi(\mathbf X_j)\big\Vert_2\bigg) -\gamma \bigg)\, \mathbf X_j \Bigg],\\
[\mathbf B_2]_i &= \mathbf X_i + h \, \mathcal K_2 \Bigg[ \displaystyle\frac{c_{n}}{\displaystyle\sum_j \omega(\mathbf X_i, \mathbf X_j)}\, \mathlarger{\mathlarger{\sum}}_j \bigg(-2\lambda\log\bigg(\big\Vert\theta(\mathbf X_i) - \phi(\mathbf X_j)\big\Vert_2\bigg) -\gamma \bigg)\, [\mathbf B_1]_j \Bigg].
\end{aligned}\end{gather}For the case $0<s<n/2$, the kernel entry is again replaced with zero, in case the denominator happens to be zero. For the case $s=n/2$, if the term $\log \big(\Vert\theta(\mathbf X_i) - \phi(\mathbf X_j)\Vert_2\big)$ turns out to be undefined, the term is then replaced by $-\frac{\gamma}{2\lambda}$ so that it does not contribute to the overall sum.

These two operators are different from the previous two discussed in this section. While for the diffusion-like operators, the kernel is integrated with the term $\mathbf X_j - \mathbf X_i$ (and $[\mathbf B_1]_j - [\mathbf B_1]_i$), for these two operators, we have the term $\mathbf X_j$ (and $[\mathbf B_1]_j$) only. This is reflected in the performance of the overall network, as we will see in Section \ref{sec:benchmark}. The computations and forward propagations of the tensors for these two nonlocal blocks with and without subsampling are treated in a similar fashion as discussed before (see Figure SM2 in the supplementary material).

The PIDE-based nonlocal blocks discussed here have several advantages over other layers in the neural network. Firstly, it is clear that, for all the operators introduced here, the pixel strips' values of the nonlocal block output depend on each pixel strip of the input tensor. This block is also different from a fully-connected layer because it computes activations based on relationships between different regions of the image, and these long-range dependencies are a function of the input data. This is different in a fully-connected layer, where the relationship is established using trainable weights. Secondly, a fully-connected layer can usually only be added at the end of the network due to computational cost reasons, and the local information is lost by flattening the image. On the other hand, the nonlocal block can be added anywhere in the network, and it preserves the 2D structure of images. Besides, the output and input shapes of this implementation are the same. Therefore, this block can be plugged into any neural network architecture associated with computer vision to accelerate the communication of information across pixels.

Note furthermore that the analogy to a Verlet time-discretization of a PIDE system can again easily be deduced as in Section \ref{sec:implement}. More specifically, the $\operatorname{Diff}$ operator in \eqref{eq:PIDEsystem} and its implicit application in \eqref{eq:nldes_revisited} just need to be substituted by the identity operator to obtain the corresponding systems for this case.

\subsection{Regularization}\label{sec:reg}
The \emph{weight-decay} or $L_2$ regularization will be used in our experiments. It penalizes large weights in the hidden layers of the network and is equivalent to the well-known Tikhonov regularization. Let $\mathbf K$ denote some (convolutional) weights of the network, then the $L_2$ regularizer is given by \begin{equation}
R_1(\mathbf K) = \frac{\alpha_1}{2} \Vert \mathbf K\Vert_F^2,
\end{equation}where $\Vert \cdot \Vert_F$ represents the Frobenius norm, and $\alpha_1$ is the regularization hyperparameter that controls the trade-off between larger and smaller trainable weights (see Section SM4).

For a stable forward propagation, we would ideally want the weight matrices $\mathbf K(t)$ to change smoothly with time, i.e. we would like convolution weights to be piecewise smooth in time. To this end, as suggested in \cite{rev}, we use \emph{weight smoothness decay} in combination with $L_2$ regularization. The smooth weight decay can be written as\begin{equation}
R_2(\mathbf K) = \alpha_2 \, h \sum_{j=1}^{N-1}\sum_{k=1}^2 \bigg\Vert \frac{\mathbf K_{j,k} - \mathbf K_{j+1,k}}{h}\bigg\Vert_F^2 ,
\end{equation}
where $h$ is the step size, $\alpha_2$ is again a hyperparameter (see Section SM4), and $j$ represents each time step or each block of the PDE-based neural network. Here it is assumed that there are two convolution layers/weights $\mathbf K_1$ and $\mathbf K_2$ in each block/time step (in Hamiltonian blocks, for example). Because of the unique nature of nonlocal blocks, the weight smoothness decay is only applied to the normal blocks and not to the nonlocal blocks, whereas $L_2$ weight decay is applied to all the convolutional network weights. Both regularization terms are then added to the objective function \eqref{eq:invprob}, which the optimizer tries to then minimize.

\subsection{Further implementation details} \label{sec:further}
After the starting layer of $3\times 3$ convolutions, the original Hamiltonian network consists of 3 Units \cite{rev} with $m$ Hamiltonian blocks each. Average pooling with pool size 2 is used to decrease the size of the feature map between the Units, followed by $1\times 1$ convolution (with ReLU) instead of zero-padding to increase the number of channels. The convolutions within each block are all $3\times 3$ convolutions with spatial zero-padding to maintain feature map size. Further implementation details regarding the choice of hyperparameters etc. can be found in the supplementary material (Section SM4).

The nonlocal block is then added after the second block in each Unit, and this network is trained and tested on several image classification benchmark datasets, such as CIFAR-10, CIFAR-100 \cite{cifar} and STL-10 \cite{stl}. The networks are also used for the semantic segmentation task in autonomous driving. To this end, the BDD100K dataset \cite{bdd} is used, which is a large-scale dataset of visual driving scenes. The resolution of each image in this dataset is $1280\times 720$. To demonstrate the performance for the segmentation task, we use small networks, and therefore we resize the images to a resolution of $160 \times 90$. This is done by removing every second row and column from each image and performing the same operation iteratively on the resulting image.

For training with CIFAR-10 and CIFAR-100, the mini-batch size is kept at 100. The per-pixel mean from the input image is subtracted before training \cite{kriz2012}, and the means of the training data are used to perform per-pixel mean subtraction on the test data. Common data augmentation techniques \cite{dataaug} are performed, such as padding 4 zeros around the image, followed by random cropping and random horizontal flipping. This makes the network more robust and helps it to generalize better. At the end of the network, before the fully-connected layer, we have an average pooling layer with a pool size of 2 (Figure \ref{fig:hamnet}). For subsampling within the nonlocal block, max-pooling is performed with a pool size of 2, i.e. $2\times 2$ patches of the image are used to compute the affinity kernel $\omega$.

For the STL-10 dataset, the mini-batch size is 50. The same preprocessing and data augmentation techniques are used but with a padding of 12 zeros around the image before cropping, instead of 4, because the images in the STL-10 dataset are significantly larger. At the end of the network, before the fully-connected layer, we have an average pooling layer with pool size 8. For subsampling within the nonlocal block, max-pooling is performed with a pool size of 4 in this case.

For the segmentation task with the BDD100K dataset, the mini-batch is of size 8. This value is intentionally kept small to avoid a large memory footprint due to the kernel computation and the relatively high image resolution. For the data augmentation, the per-pixel mean subtraction is performed, and then the image is randomly flipped horizontally. In a usual neural network for semantic segmentation tasks, the image is subsampled several times and then upsampled again to get a high-dimensional output. We use a simple architectural design here, i.e. the pooling layers that are shown in Figure \ref{fig:hamnet} are left out. This way, the spatial dimensions of the image are maintained. At the end of the network, instead of a fully-connected layer, the output of the last unit is passed through a $1\times 1$ convolutional layer with 20 filters, which gives us the prediction of each of the pixels in the image. For subsampling within the nonlocal block, max-pooling is performed with a pool size of 3. 
\section{Numerical Experiments}

We will denote the networks with the nonlocal blocks as nonlocal Hamiltonian networks, and the one without the nonlocality is just called Hamiltonian network. Each experiment is run five times with different random seeds, and the median test accuracy is reported in each case. In \cite{rev}, the smallest Hamiltonian network proposed is a network with 74 layers, with 6 Hamiltonian blocks in each of the three Units (6-6-6). The initial convolutional layer has 32 filters, and the number of channels in each Unit is \{32, 64, 112\}. Such architectures with three Units have been used in ResNets and its variants, and have become a common blueprint for a well-performing CNN. We stick to the simple (6-6-6) architecture for our experiments. The last Unit intentionally has 112 channels to make sure that the resulting network is comparable to the baseline ResNet-44, in terms of the number of trainable parameters.

\subsection{Training on benchmark datasets} \label{sec:benchmark}
Firstly, the nonlocal Hamiltonian networks are tested on the benchmark datasets CIFAR-10, CIFAR-100 and STL-10. The original Hamiltonian network, ResNet-44 and PreResNet-20 \cite{he2016b} are used as baseline networks for comparison and are trained on each of the three datasets. The main results with the test accuracies are shown in Table \ref{tab:benacc}. The best result for each benchmark dataset is marked in boldface.

\renewcommand{\arraystretch}{1.3}
\begin{table}[h!]
\centering
\caption{Test accuracies on benchmark datasets for different nonlocal Hamiltonian networks.}
\resizebox{0.95\textwidth}{!}{%
\begin{tabular}{|l|c|c|c|c|c|c|}
\hline
\multicolumn{1}{|c|}{\multirow{2}{*}{\textbf{Network}}} & \multicolumn{2}{c|}{\textbf{CIFAR-10}}   & \multicolumn{2}{c|}{\textbf{CIFAR-100}}  & \multicolumn{2}{c|}{\textbf{STL-10}}     \\ \cline{2-7} 
 & \textbf{\small\hspace{-0.15cm}Params (M)\hspace{-0.15cm}} & \textbf{\small Acc. (\%)} & \textbf{\small\hspace{-0.15cm}Params (M)\hspace{-0.15cm}} & \textbf{\small Acc. (\%)} & \textbf{\small\hspace{-0.15cm}Params (M)\hspace{-0.15cm}} & \textbf{\small Acc. (\%)} \\\hhline{|=|=|=|=|=|=|=|}
ResNet-44 & $0.66$ & $92.64$ & $0.66$ & $69.51$ & $0.66 $ & $75.79$ \\ 
PreResNet-20 & $0.57$ & $92.35$ & $0.59$ & $71.07$ & $0.57$ & $77.39$ \\
Hamiltonian-74 & $0.50$ & $92.75$ & $0.67$ & $70.38$ & $0.50$ & $79.95$ \\ \hline
Nonlocal diffusion  $\mathcal L$ & \multirow{4}{*}{$0.56$} & $\mathbf{93.27}$ & \multirow{4}{*}{$0.72$} & $71.81$ & \multirow{4}{*}{$0.55$} & $\mathbf{82.62}$   \\ 
Pseudo-differential $(-\Delta)^{1/2}$ & & $93.21$ & & $\mathbf{71.84}$ & & $81.88$       \\ 
Pseudo-differential $(-\Delta)^{-1/2}$ & & $93.08$ & & $71.24$ &  & $81.56$      \\ 
Pseudo-differential $(-\Delta)^{-1}$ & & $92.88$ & & $71.25$  & & $80.73$   \\ \hline
\end{tabular}
}
\label{tab:benacc}
\end{table}\renewcommand{\arraystretch}{1}

The figures in Table \ref{tab:benacc} show that all of the nonlocal Hamiltonian networks perform better on the benchmark datasets than the baseline networks. They also have fewer parameters than the baseline networks of ResNet and PreResNet networks (for CIFAR-10 and STL-10). The nonlocal diffusion $\mathcal L$ network performs the best, closely followed by the pseudo-differential $(-\Delta)^{1/2}$ network, which can be seen as a special case of the nonlocal diffusion operator, as pointed out in Section \ref{sec:fraclap}. The inverse Laplacian operator $(-\Delta)^{-1}$ with the $\log$ kernel performs the worst, but it nevertheless outperforms the baseline networks. Besides, the accuracies of our networks are rather comparable to that of ResNet-56 and ResNet-110 and not to ResNet-44 (see \cite{he2016}). This reiterates the fact that the nonlocal connections and their larger receptive fields somewhat compress the ResNet-like architectures. These connections partially alleviate the necessity to have very deep networks, and hence, we can save training time by training shallower networks with nonlocal blocks in them.

The nonlocal diffusion network performs better than the one proposed in \cite{taolocal}, which did not have any skip connections in the nonlocal block. This suggests that introducing skip connections within the nonlocal block makes sure that the network learns only the update (or the residual) made to the identity function, and hence avoids any impedance when the information travels through the nonlocal block.

\begin{figure}[h!]
\centering
\includegraphics[scale=0.28]{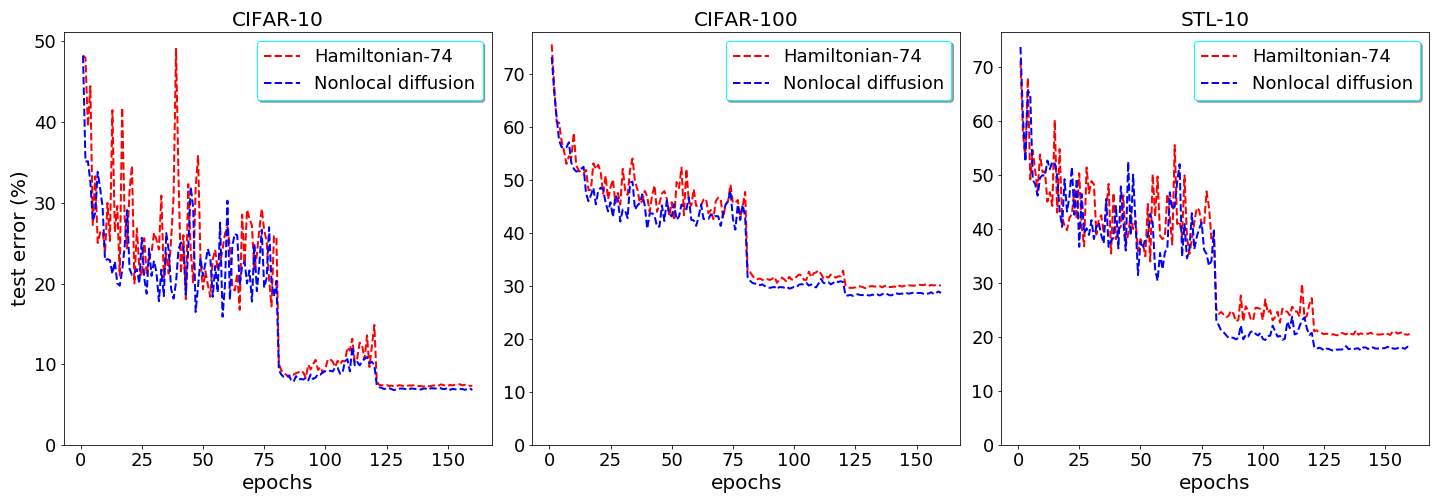}
\caption{Test accuracy curves for the nonlocal diffusion $\mathcal L$ network in comparison to the original Hamiltonian network.}
\label{fig:typ1_curve}
\end{figure}

The nonlocal diffusion $\mathcal L$ network and the pseudo-differential $(-\Delta)^{1/2}$ network perform comparatively better than the other two proposed neural networks. This is possibly because of the difference term $[u(\mathbf x) - u(\mathbf y)]$ (and [$u(\mathbf y)- u(\mathbf x)$]). When computing each pixel in the output feature map, this term compares and computes the relative differences between the neighboring pixel values of the input feature map, whereas the operator $(-\Delta)^{-1/2}$ just computes a weighted sum of $u(\mathbf y)$ (see Definition \ref{def:invfraclap}). Also, we see the biggest increase in accuracy for the STL-10 dataset. This is due to the fact that the images in this dataset are relatively larger than the ones in CIFAR-10/100, and therefore the nonlocality plays a more vital role in widening the receptive field of the network, which enables direct communication between pixels on opposite ends of the image. As an example, the test accuracy curves (till the 160th epoch) for the nonlocal diffusion network with each of the three datasets are shown in Figure \ref{fig:typ1_curve}. The Hamiltonian-74 network is used for comparison since it has the best performance out of all the three baseline networks.

The nonlocal Hamiltonian networks can actually achieve better results for the STL-10 dataset. With a larger step size $h$, around $84\%$-$84.5\%$ test accuracy on STL-10 can be attained. For example, the pseudo-differential $(-\Delta)^{1/2}$ network was trained with a higher value of $h$, namely $h=0.15$, and it achieved a median test accuracy of $84.12\%$. The value of $h$ is intentionally kept low because, as pointed out in \cite{giantleap}, a small $h$ encourages the model to have smaller weights. A model with larger weights tends to suffer from overfitting. Such models are also vulnerable to adversarial attacks.

It has been seen that although the pseudo-differential $(-\Delta)^{1/2}$ network performed slightly worse than the nonlocal diffusion network on the benchmark datasets (Section \ref{sec:benchmark}), it is at times more robust to noise and perturbations \cite{perturb2} than the nonlocal diffusion and Hamiltonian networks. This shows that a higher test accuracy of a network does not automatically mean more robustness to noise. The experiments have also shown that the nonlocal Hamiltonian networks, on average, need lesser training data to learn the features in the dataset and to achieve a certain generalization power. 

\subsection{Multi-stage nonlocal blocks} \label{sec:msnb}
The discretizations of the nonlocal operators relate to a 2-stage computation (Figure \ref{fig:twostage}). To see the effect of the number of stages in the nonlocal block, the nonlocal diffusion network with four stages is taken as an example and is trained on the datasets. The results are shown in Table \ref{tab:benacc3}. 
\renewcommand{\arraystretch}{1.3}
\begin{table}[h!]
\centering
\caption{Test accuracies on benchmark datasets for the nonlocal diffusion $\mathcal L$ network with different number of stages in the nonlocal block.}
\scalebox{0.8}{
\begin{tabular}{|c|c|c|c|c|c|c|}
\hline
\multirow{2}{*}{\textbf{\# Stages}} & \multicolumn{2}{c|}{\textbf{CIFAR-10}}   & \multicolumn{2}{c|}{\textbf{CIFAR-100}}  & \multicolumn{2}{c|}{\textbf{STL-10}}     \\ \cline{2-7} 
 & \textbf{\small\hspace{-0.15cm}Params (M)\hspace{-0.15cm}} & \textbf{\small Acc. (\%)} & \textbf{\small\hspace{-0.15cm}Params (M)\hspace{-0.15cm}} & \textbf{\small Acc. (\%)} & \textbf{\small\hspace{-0.15cm}Params (M)\hspace{-0.15cm}} & \textbf{\small Acc. (\%)} \\ \hhline{|=|=|=|=|=|=|=|}
2 & 0.56 & 93.27 & 0.72 & 71.81 & 0.55 & 82.62  \\ 
4 & 0.59 & 93.09 & 0.76 & 71.25 & 0.59 & 81.36 \\ \hline
\end{tabular}%
}
\label{tab:benacc3}
\end{table}\renewcommand{\arraystretch}{1}

The figures indicate that too many stages in the nonlocal block damp the signal in the network, and hence, the network performs worse or at least does not perform any better than the network with a 2-stage nonlocal block. This is in line with the observations made in \cite{taolocal}. Operators similar to the inverse fractional Laplacian also suffer from instabilities if the number of stages is increased (see \cite{taolocal}). Therefore, although a nonlocal operation of this kind improves performance when the number of stages is kept low, overusing the nonlocal operation can lead to a lossy network.

\subsection{Semantic segmentation on BDD100K} \label{sec:segment}
The aim of semantic segmentation is to predict the class of each pixel and thereby partition the image into different image classes or objects. This gives us an idea of where the object is present in the image and which pixels belong to it. This is of paramount importance, for example, for self-driving cars because the position and type of the object in front are necessary inputs to any self-driving algorithm.

\begin{figure}[h!]
\centering
\includegraphics[scale=0.33]{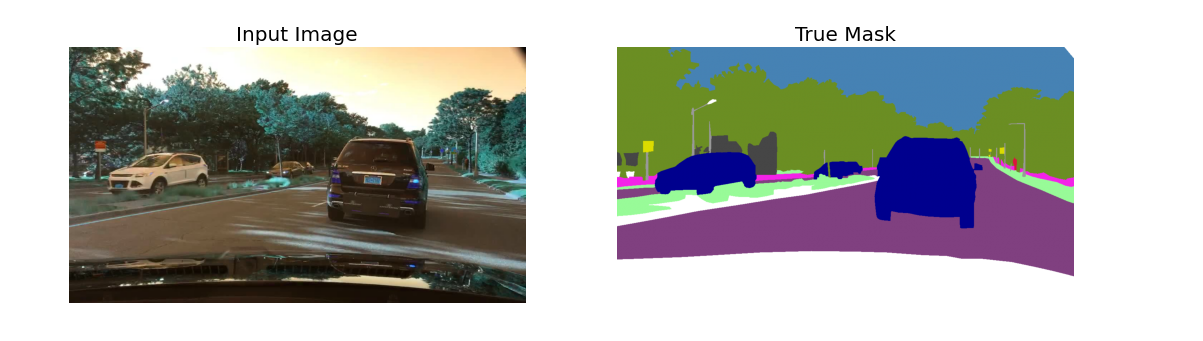}
\caption{Input images and corresponding labels from Berkeley DeepDrive (BDD100K) dataset for the semantic segmentation task \cite{bdd}.}
\label{fig:bdd}
\end{figure}

The nonlocal Hamiltonian networks are used to segment images from the BDD100K dataset \cite{bdd}, which consists of images with everyday driving scenarios. An example of an image, along with its segmentation mask, is provided in Figure \ref{fig:bdd}. Our aim here is to show the increase in performance due to the presence of nonlocal connections and not to challenge the state-of-the-art networks. Therefore, we employ simple architectures with no pooling layers to maintain the spatial dimension of the feature maps, and we work with smaller image resolutions, as mentioned in Section \ref{sec:further}.

\renewcommand{\arraystretch}{1.3}
\begin{table}
\centering
\caption{Pixel accuracies and meanIoU for the semantic segmentation task on the BDD100K dataset using the nonlocal Hamiltonian networks.}
\scalebox{0.8}{
\begin{tabular}{|l|c|c|c|}
\hline
\multicolumn{1}{|c|}{\multirow{2}{*}{\textbf{Network}}} & \multicolumn{3}{c|}{\textbf{BDD100K}}     \\ \cline{2-4} 
 & \textbf{\small\hspace{-0.15cm}Params (M)\hspace{-0.15cm}} & \textbf{\small P. Acc. (\%)} & \textbf{\small mIoU (\%)} \\\hhline{|=|=|=|=|}
ResNet-44 & $0.66$ & 83.50 & 63.42 \\ 
PreResNet-20 & $0.57$ & 77.86 & 52.12 \\
Hamiltonian-74 & $0.49$ & 84.47 & 64.40 \\ \hline
Nonlocal diffusion  $\mathcal L$ & \multirow{4}{*}{$0.54$} & 84.80 & $\mathbf{67.01}$  \\ 
Pseudo-differential $(-\Delta)^{1/2}$ & & $\mathbf{85.24}$ & 66.71  \\ 
Pseudo-differential $(-\Delta)^{-1/2}$ & & 85.21 & 66.52  \\ 
Pseudo-differential $(-\Delta)^{-1}$ & & 84.95 & 66.26  \\ \hline
\end{tabular}
}
\label{tab:seg}
\end{table}\renewcommand{\arraystretch}{1}

\begin{figure}[h!]
\centering
\includegraphics[scale=0.3]{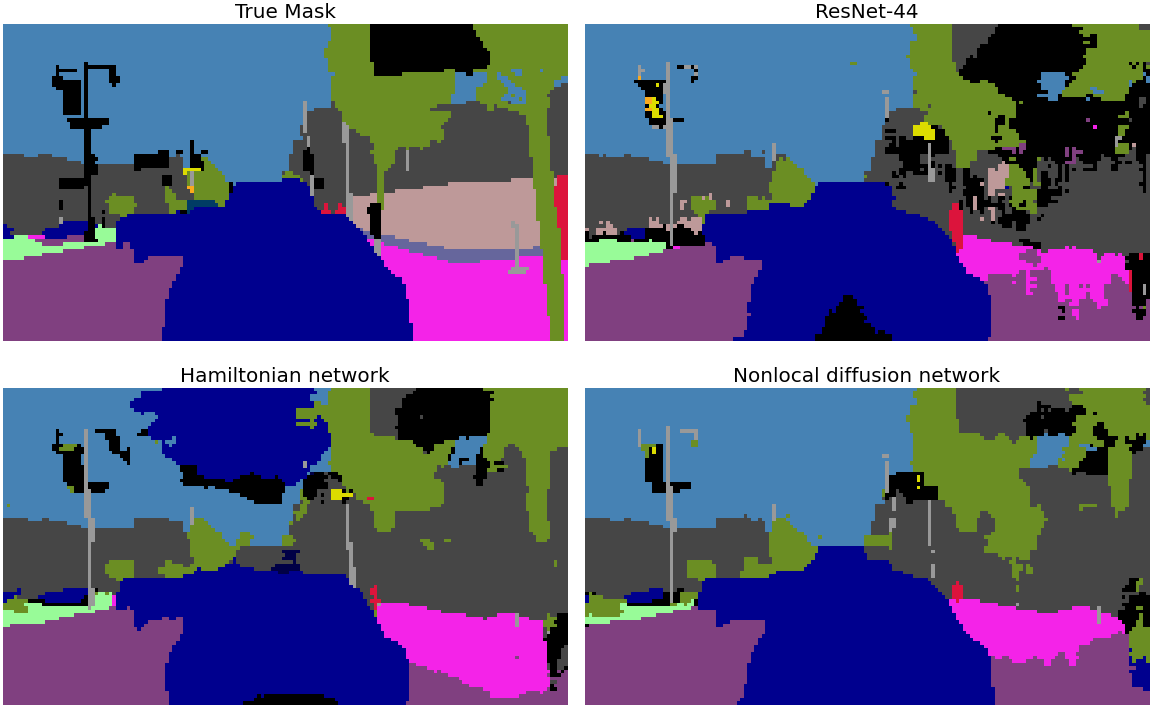}
\caption{The true labels/masks of a test image from the BDD100K dataset and the predictions made by the different networks.}
\label{fig:bdd2}
\end{figure}

To evaluate the performance, two metrics will be used. The first one is the pixel accuracy. The second metric that we use is the Intersection-over-Union (IoU) metric (or the \emph{Jaccard index}). This is calculated on a per-class basis, and the mean is then taken over all the classes, which gives us the meanIoU (mIoU) metric. The results for the proposed neural networks can be seen in Table \ref{tab:seg}. When we consider the pixel accuracy and the mIoU metric, the nonlocal Hamiltonian networks perform better than the baseline networks. The nonlocal Hamiltonian networks' metric values are very close to each other. We see that the nonlocal Hamiltonian networks have a performance gain of around 2--2.5 mIoU percentage points when we compare it to the original Hamiltonian network and around 3--3.5 mIoU percentage points when we compare it to ResNet-44. To make sure that the networks actually learn to segment the images from driving scenes into semantically meaningful parts, we look at the predictions made by the networks. Figure \ref{fig:bdd2} shows the true label of a test image (with a taxi) and the different predictions made by some of the networks discussed above.
The results discussed here can be improved by performing segmentation using encoder-decoder models like U-Net \cite{unet} or data augmentation techniques, which we did deliberately not use here to completely focus on the comparison of the networks architectures themselves.

\section{Computational cost and forward propagation stability}
\subsection{Computational cost of the nonlocal blocks} \label{sec:cost}
\renewcommand{\arraystretch}{1.3}
\begin{table}
\centering
\caption{Number of floating-point operations for each of the networks when trained on the image classification benchmark datasets.}
\scalebox{0.8}{
\begin{tabular}{|l|c|c|c|}
\hline
\multicolumn{1}{|c|}{\multirow{2}{*}{\textbf{Network}}} & \textbf{CIFAR-10}  & \textbf{CIFAR-100} & \textbf{STL-10}    \\ \cline{2-4} 
\multicolumn{1}{|c|}{} & \textbf{\small FLOPs (M)} & \textbf{\small FLOPs (M)} & \textbf{\small FLOPs (M)} \\ \hhline{|=|=|=|=|}
ResNet-44 & 98.3 & 98.3 & 879.5   \\
Hamiltonian-74 & 159.6 & 159.9 & 1432.5       \\ \hline
Nonlocal diffusion  $\mathcal L$ & 192.9 & 193.2 & 2003.7   \\
Pseudo-differential $(-\Delta)^{1/2}$ & 193.4 & 193.7 & 2012.5 \\
Pseudo-differential $(-\Delta)^{-1/2}$ & 193.2 & 193.5 & 2011.0  \\
Pseudo-differential $(-\Delta)^{-1}$ & 193.6 & 193.9 & 2019.5 \\ \hline
\end{tabular}%
}
\label{tab:flops}
\end{table}\renewcommand{\arraystretch}{1}

The nonlocal blocks come with a computational cost that needs to be taken into consideration while inserting them into any neural network. The number of FLOPs (multiply-adds) associated with the proposed networks for the image classification task are shown in Table \ref{tab:flops}. To ensure the stability of the forward propagation, we can observe that the Hamiltonian-74 network has roughly 1.5 times more FLOPs than the traditional ResNet-44. Also, the ResNets generally have only two convolutional layers in a Residual block. In comparison, Hamiltonian blocks have 2 convolutions and 2 transposed convolutions, which explains the higher number of FLOPs. The nonlocal blocks by themselves add only little extra floating-point operations for CIFAR-10 and CIFAR-100 but nevertheless help the networks achieve better accuracy on the benchmark datasets. For the effect of subsampling on the computational costs for STL-10 computations, see section SM5 in the supplementary material.

\subsection{Stability of the PIDE-based forward propagation} \label{sec:eigsym}
It is crucial that the newly introduced PIDE-based nonlocal block does not introduce any instabilities in the forward propagation. If it did, then it would nullify the advantage of the PDE-based Hamiltonian network having the stability of the forward propagation. The spectral norms of the weight matrices have a major role to play in this regard. To investigate this, we look at the spectral nature of the weights in the nonlocal block. 

\begin{figure}[h!]
\centering
\includegraphics[scale=0.3]{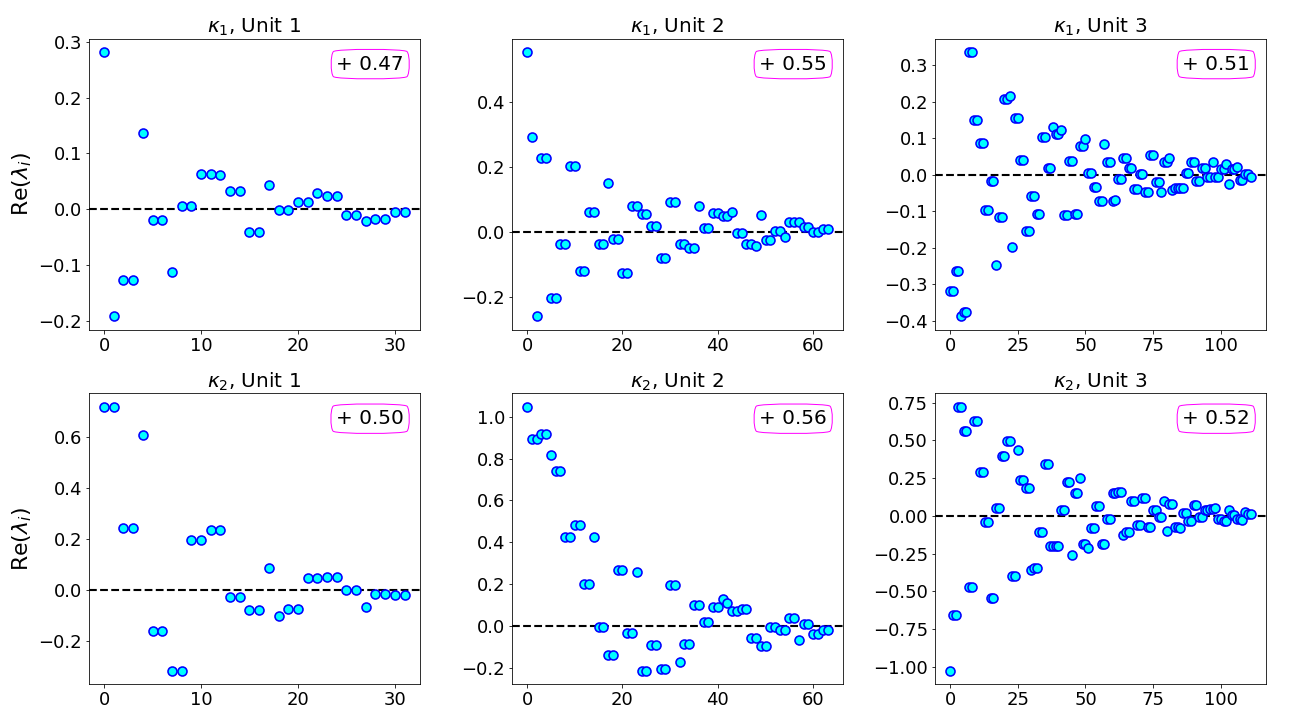}
\caption{Real parts of the eigenvalues of weight matrices $\mathcal K_1$, $\mathcal K_2$ in the nonlocal block placed in each Unit of the nonlocal diffusion $\mathcal L$ network while training on STL-10.}
\label{fig:eigunit2}
\end{figure}

Other than the embeddings $\theta$ and $\phi$, the nonlocal block has two $1\times 1$ convolutional layers $\mathcal K_1$ and $\mathcal K_2$ for the two stages. For the four nonlocal Hamiltonian networks used in Section \ref{sec:benchmark}, the weights of the convolution layers have the dimensions $\mathcal K_{1,2} \in \mathbb{R}^{32\times 32}$, $\mathcal K_{1,2} \in \mathbb{R}^{64\times 64}$, $\mathcal K_{1,2} \in \mathbb{R}^{112\times 112}$, and they belong to the three Units of the network, respectively. The outputs of these transformations are the feature maps that are used later on in other blocks of the network. Thus, we look at the eigenvalues of these matrices to see how the features are transformed by these trainable weights. Figure \ref{fig:eigunit2} shows the real parts of the eigenvalues of the weight matrices $\mathcal K_1$ and $\mathcal K_2$ in each Unit of the nonlocal diffusion network while training on STL-10. The real parts of the eigenvalues for the other nonlocal Hamiltonian networks can be found in the supplementary material.

The plots show that the real parts of the eigenvalues of the trainable weights are mostly very close to zero. The \enquote{$+$} sign in each plot indicates the fraction of eigenvalues that have positive real parts. In the ideal case, this value should stay close to 0.5. Too many eigenvalues with positive real parts would lead to amplification of the signal resulting in instabilities in the forward propagation. On the other hand, too many eigenvalues with negative real parts would lead to a lossy network. The plots for the nonlocal diffusion network are pretty symmetric. The plots for the pseudo-differential networks $(-\Delta)^{1/2}$, $(-\Delta)^{-1/2}$ and $(-\Delta)^{-1}$ (Figures SM3, SM4, and SM5 in the supplementary material) show that the real parts are slightly more asymmetrically distributed, which partly explains their relatively worse performance.

There is another way of looking at the properties of the transformations carried out by the trainable weight matrices $\mathcal K_1$ and $\mathcal K_2$. Let $\mathcal K \in \{\mathcal K_1, \mathcal K_2\}$. Then the amplification or damping of the features by $\mathcal K\in \mathbb{R}^{n \times n}$ is mainly determined by the associated quadratic form $\mathbf x^T \mathcal K \mathbf x \in \mathbb R$, for $\mathbf x \in \mathbb R^n$. We can decompose $\mathcal K$ into a symmetric and an antisymmetric part \[\mathcal K = \frac{\mathcal K + \mathcal K^T}{2} + \frac{\mathcal K - \mathcal K^T}{2} = \mathcal K_{s} + \mathcal K_{as}.\]Then the quadratic form of $\mathcal K$ is given by
$\mathbf x^T \mathcal K \mathbf x =  \mathbf x^T \mathcal K_s \mathbf x$, where the quadratic form of the antisymmetric matrix $\mathcal K_{as}$ is zero. Therefore, the quadratic form of the symmetric part of $\mathcal K$, i.e. the quadratic form of $\mathcal K_s$, determines the spectral properties of the transformations. Since $\mathcal K_s$ is symmetric, the eigenvalues are all real.

\begin{figure}[h!]
\centering
\includegraphics[scale=0.3]{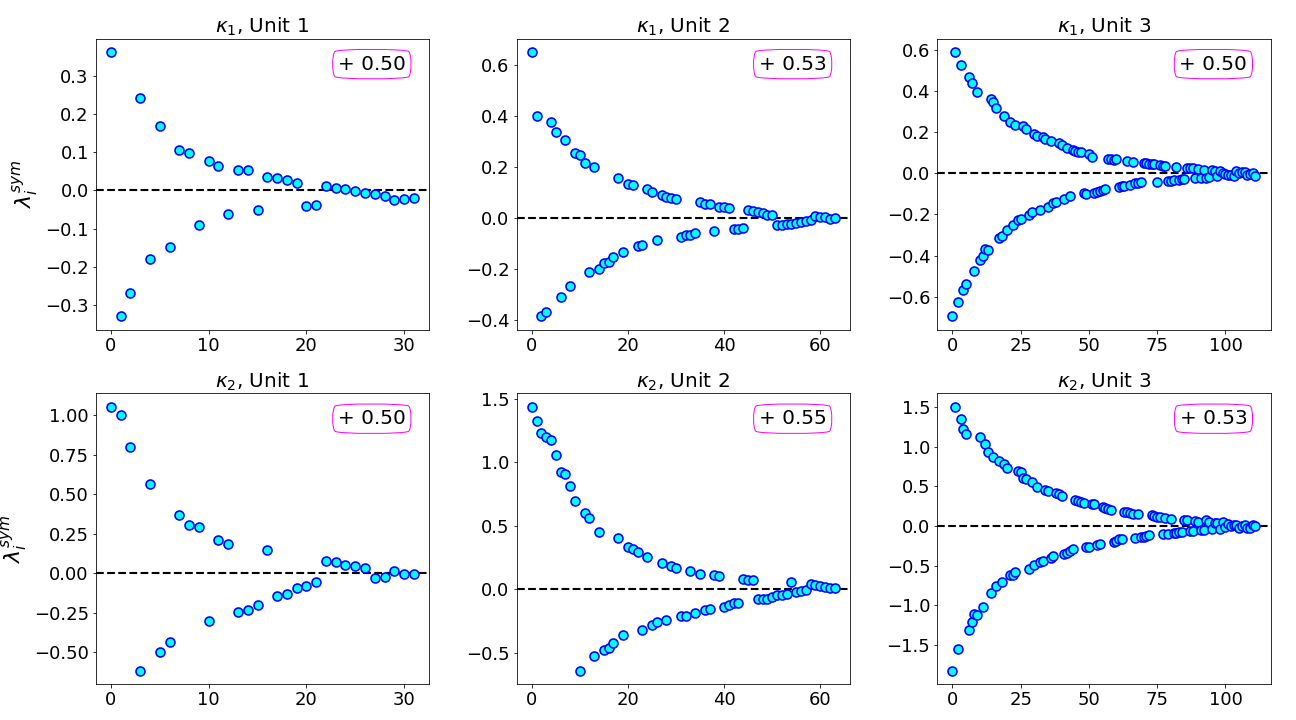}
\caption{Eigenvalues of the symmetric parts of weight matrices $\mathcal K_1$, $\mathcal K_2$ in the nonlocal block placed in each Unit of the nonlocal diffusion $\mathcal L$ network while training on STL-10.}
\label{fig:sym2}
\end{figure}

Figure \ref{fig:sym2} shows the eigenvalues of the symmetric parts of the weight matrices $\mathcal K_1$ and $\mathcal K_2$ in each Unit in the nonlocal diffusion network while training on STL-10. The eigenvalues of the symmetric parts for the other nonlocal Hamiltonian networks can be found in the supplementary material (Figures SM6, SM7, and SM8). The \enquote{$+$} sign in each plot indicates the fraction of positive eigenvalues. The plots show that most of the eigenvalues are well-bounded and symmetric, unlike in the case of the network suggested in \cite{helocal}, where the eigenvalues of the symmetric part are significantly larger when multiple stages are added to the nonlocal block, as shown in \cite{taolocal}. This well-boundedness of the eigenvalues makes the task of the optimizer easier, which leads to faster convergence.

\section{Conclusion}
\ifx
To improve a network's performance, the recent trend is to have deeper or/and wider networks. Instead, this paper proposes a network design involving nonlocal blocks that introduce nonlocal connectivities between pairs of regions of the image that accelerate the communication between far-away pixels and thereby widen the receptive field of the network. Due to an appropriate combination of local and nonlocal information, the network gets a more comprehensive view of each example during training. As a consequence, this helps the network to generalize better on unseen test examples. These nonlocal operators, although dense, are discretized in an efficient way using $1\times 1$ convolutions and pooling such that we have a relatively low extra computational cost and very little increase in the number of trainable parameters. Similar to ResNets, the nonlocal blocks contain skip connections, which make sure that their presence does not hinder the flow of information in the network.\fi

The proposed neural networks achieve better accuracy than the original Hamiltonian model and the widely-used state-of-the-art ResNets. Due to an appropriate combination of local and nonlocal information, the network gets a more comprehensive view of each example during training. The nonlocal diffusion $\mathcal L$ network and the pseudo-differential $(-\Delta)^{s}$ network perform slightly better than the other two proposed networks. Moreover, the PIDE-based nonlocal block maintains the 2D structure of the images and has the same output shape as the input shape. Hence, they can be seamlessly inserted into any network. This can be useful if a neural network is to be trained and tested on a dataset that warrants nonlocal connections between the regions of the image. Using spectral properties of the transformation matrices, we showed that the nonlocal blocks do not destabilize the original Hamiltonian network in any way, and consequently, the stability of the forward propagation is still preserved.

\ifx
For future work, one can concentrate on improving the training time of these networks. While using the nonlocal operators, often the mini-batch size is reduced to keep the memory footprint in check, although this has an adverse effect on the training time. Besides, the nonlocal diffusion operator $\mathcal L$ can be seen as a global weighted graph Laplacian. So it would be interesting to see if it finds any use in Graph Convolutional Networks (GCNs) \cite{graphcnn}. Moreover, the nonlocal nature could be tested in combination with implicit discretizations \cite{imex} of the underlying PDE/ODE of the neural network. Implicit discretizations are actually known to broaden the receptive field of the network \cite{imex}. Hence, a connection to nonlocal operators is an interesting topic for further research.

The interpretations of ResNets and very deep networks, in general, are still an on-going research topic and hopefully, the dynamical system view coupled with spatially nonlocal operators will shed some light on the understanding and interpretability of these networks. This paper hopes to contribute to the research on \emph{structural diversity} by providing alternative network designs, which could inspire further work in this direction in the future.
\fi
\section*{Acknowledgments}
Michael Griebel was supported by the {\em Hausdorff Center for Mathematics} in Bonn, funded by the Deutsche Forschungsgemeinschaft (DFG, German Research Foundation) under Germany's Excellence Strategy  - EXC-2047/1 - 390685813 and the CRC 1060 {\em The Mathematics of Emergent Effects} of the Deutsche Forschungsgemeinschaft.

\bibliographystyle{siamplain}
\bibliography{references}

\end{document}


\maketitle
\section{Pseudo-differential operators}\label{sec:pdo}
\begin{definition}
Let $\Omega \subset \mathbb R^n$ be open, $0\leq\rho\leq 1$, $0\leq \delta\leq 1$, $m\in \mathbb R$, $n\in \mathbb N$, $n\geq 1$. Then the space of symbols of order $m$ and of type $(\rho, \delta)$, denoted by $S^{m}_{\rho,\delta}$, is the space of all $p\in C^\infty(\Omega\times \mathbb R^n)$ such that, for all compact sets $K\subset \Omega$ and all multi-indices  $\alpha, \beta \in \mathbb N^n$, there is a constant $C_{K, \alpha,\beta}$ such that \[
\vert \, \partial_{\mathbf x}^\alpha \, \partial_{\mathbf \xi}^\beta \, p(\mathbf x, \xi)\vert \leq C_{K,\alpha,\beta} (1+\vert \xi\vert)^{m-\rho\vert\beta\vert+\delta\vert\alpha\vert},\]where $\mathbf x\in K$, $\xi \in \mathbb R^n$. 
\end{definition}
\begin{definition}
Let $v\in \mathcal S(\mathbb R^n)$, $p \in S^{m}_{\rho,\delta}$. A pseudo-differential operator $P(\mathbf x, D)$ on $\mathbb R^n$ with symbol $p(\mathbf x, \xi)$ is defined as
\begin{equation}
P(\mathbf x, D) v(\mathbf x) := \int\limits_{\mathbb R^n} p(\mathbf x, \xi) \, \hat v(\xi)\, e^{2\pi i\mathbf x\cdot \xi} \, d\xi.
\end{equation}  
\end{definition}

\section{Addressing the singularity}\label{sec:singularity}
The kernel under the integral in equation (3.5) has a singularity when $\mathbf x = \mathbf y$, but for $v\in \mathcal S(\mathbb R^n)$, we can write

\begin{align*}
\int\limits_{\mathbb R^n} \frac{[v(\mathbf x) - v(\mathbf y)]}{\vert \mathbf x - \mathbf y\vert^{n+2s}}  \, d\mathbf y & \leq C  \int\limits_{B_R(\mathbf x)} \frac{\vert \mathbf x -\mathbf y \vert}{\vert \mathbf x - \mathbf y\vert^{n+2s}} \, d\mathbf y + \Vert v \Vert_{L^\infty(\mathbb R^n)} \int\limits_{\mathbb R^n \setminus B_R(\mathbf x)} \frac{1}{\vert \mathbf x - \mathbf y\vert^{n+2s}} \, d\mathbf y\\
& = C \bigg( \int\limits_{B_R(\mathbf x)} \frac{1}{\vert \mathbf x - \mathbf y\vert^{n+2s-1}} \, d\mathbf y + \int\limits_{\mathbb R^n\setminus B_R(\mathbf x)} \frac{1}{\vert \mathbf x - \mathbf y\vert^{n+2s}} \, d\mathbf y\bigg)\\
&= C \bigg( \int \limits_0^R \frac{1}{\vert \mathbf r \vert^{2s}} \, d\mathbf r + \int\limits_R^{\infty} \frac{1}{\vert \mathbf r\vert^{2s+1}} \, d \mathbf r\bigg) < +\infty.
\end{align*} $C$ is a positive constant depending on $n$ and $\Vert v \Vert_{L^\infty}$. The integral shown above is finite only for $0<s<1/2$, but we can make the integral in equation (3.5) well-defined for $0<s<1$. 

It can be shown \cite{hitch2} that the fractional Laplacian operator given by (3.5) can also be written as $(-\Delta)^s v(\mathbf x) := \frac{c_{n,s}}{2}  \int\limits_{\mathbb R^n} \frac{[2v(\mathbf x) - v(\mathbf x+\mathbf y)-v(\mathbf x-\mathbf y)]}{\vert\mathbf y\vert^{n+2s}} \, d\mathbf y$. Using this reformulation of the fractional Laplacian and the second-order Taylor expansion of $u$ and assuming $v$ to be smooth, the integral term \[\int\limits_{\mathbb R^n} \frac{[2v(\mathbf x) - v(\mathbf x+\mathbf y)-v(\mathbf x-\mathbf y)]}{\vert\mathbf y\vert^{n+2s}} \leq \frac{\Vert D^2v\Vert_{L^\infty}}{\vert \mathbf y\vert^{n+2s-2}}\]is integrable at the origin for any $0<s<1$. Thus, we can remove $ P.V.$ from the integral as long as $v\in \mathcal S(\mathbb{R}^n)$. The idea is that, near $\mathbf x$, $[v(\mathbf x)-v(\mathbf y)]$ has the approximation $\nabla v(\mathbf x)\cdot (\mathbf x-\mathbf y)$, and the term under the integral is of the form $\frac{\nabla v(\mathbf x)\cdot (\mathbf x-\mathbf y)}{\vert \mathbf x - \mathbf y\vert^{n+2s}}$. This term is odd with respect to $\mathbf x$, and consequently, it averages out for any $\mathbf y$ in the neighborhood of $\mathbf x$ by symmetry, and the immediate neighborhood does not contribute to the integral.

\section{Speeding up tensor computations}\label{sec:speedup}
\subsubsection*{Rearranging the two-step computation}
The affinity kernel $\omega$ is stored in a matrix form for each batch of training data. We can reorder the terms in equation (4.1) to enable faster computations of tensors. For example, a part of the term in the first equation of (4.1) can be written as 
\begin{align}
\mathlarger{\mathlarger{\sum}}_j \omega(\mathbf X_i, \mathbf X_j) (\mathbf X_j - \mathbf X_i) &= \mathlarger{\mathlarger{\sum}}_j \omega(\mathbf X_i, \mathbf X_j) \mathbf X_j \,\, - \,\, \mathlarger{\mathlarger{\sum}}_j \omega(\mathbf X_i, \mathbf X_j) \mathbf X_i\\
&= \mathlarger{\mathlarger{\sum}}_j \omega(\mathbf X_i, \mathbf X_j) \mathbf X_j \, \, - \,\,\mathbf X_i \mathlarger{\mathlarger{\sum}}_j \omega(\mathbf X_i, \mathbf X_j). \label{eq:reform}
\end{align} The second part of this term is just the sum of the $i$-th row of the matrix representing $\omega$, times the pixel strip $\mathbf X_i$. Similarly, for the second stage, we have 
\begin{equation} \label{eq:reform2}
[\mathbf B_2]_i= \cdots \bigg[ \cdots \mathlarger{\mathlarger{\sum}}_j \omega(\mathbf X_i, \mathbf X_j) [\mathbf B_1]_j \, \, - \,\,[\mathbf B_1]_i \mathlarger{\mathlarger{\sum}}_j \omega(\mathbf X_i, \mathbf X_j)\bigg].\end{equation}

\subsection*{Nonlocal diffusion operator}
\begin{figure}[h!]
\centering
  \includegraphics[page=1,scale=0.8]{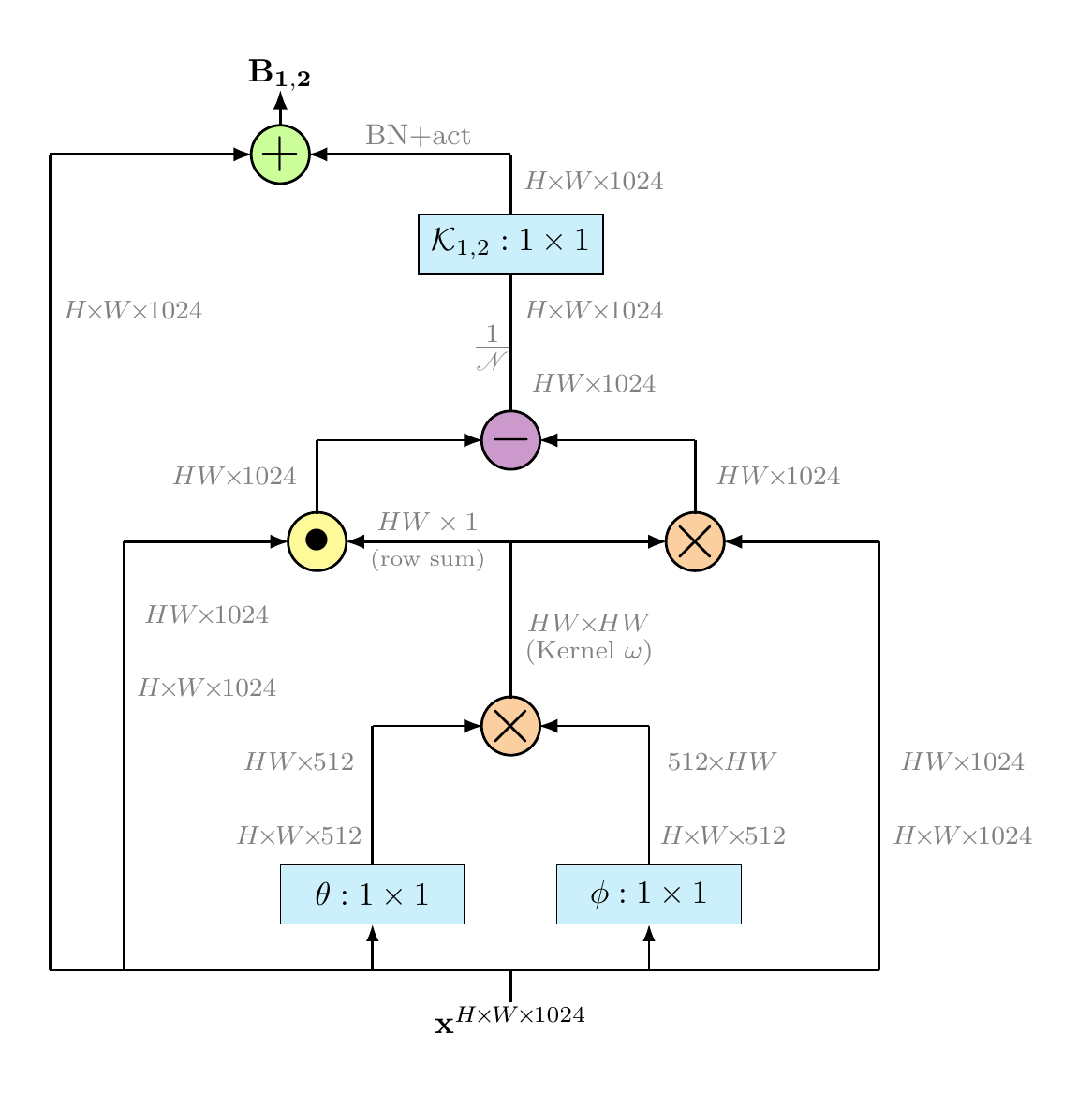}
\caption{Forward propagation of tensors in a nonlocal block with the nonlocal diffusion operator $\mathcal L$.} \label{fig:tensordiag1}
\end{figure}

Figure \ref{fig:tensordiag1} shows how the tensors are dealt with and propagated forward in the nonlocal block, where the shape of the input to the nonlocal block is assumed to be $H\times W\times 1024$, $1024$ being the number of channels of the input. $H$ and $W$ are the spatial height and width of the feature maps, respectively. The diagram only shows the computations for one stage of the nonlocal block since the other stages are just repetitions but with a pre-computed kernel $\omega$. In Figure \ref{fig:tensordiag1}, $\mathbf \otimes$ represents matrix multiplication, $\mathbf \oplus$ and $\mathbf \ominus$ denote element-wise addition and subtraction, respectively, and $\mathbf \odot$ denotes the dot product. For the dot product, each row of the tensor with shape $HW\times 1024$ is multiplied by a corresponding element from the row vector of shape $HW\times 1$. The first multiplication is used to compute the kernel $\omega$. The second multiplication represents the first term in equations \eqref{eq:reform} and \eqref{eq:reform2}, and the dot product is used to compute the second part of equations \eqref{eq:reform} and \eqref{eq:reform2}.
\subsection*{Fractional Laplacian operator $(-\Delta)^s$}
The tensors in the nonlocal block are computed similarly, as shown in Figure \ref{fig:tensordiag1}. In the case of pseudo-differential operators, the $\mathbf \otimes$ symbol for the kernel computation represents pair-wise distance computations. The other $\mathbf \otimes$ symbol in the figures still stands for matrix multiplication as before. The only difference here is that we compute $(\mathbf X_i - \mathbf X_j)$ (and $([\mathbf B_1]_i - [\mathbf B_1]_j)$), instead of $(\mathbf X_j - \mathbf X_i)$ (and $([\mathbf B_1]_j - [\mathbf B_1]_i)$), to be in line with the definitions of the two operators. However, this does not make any difference to the learning problem.
\begin{figure}[h!]
\centering
  \includegraphics[page=1,scale=0.8]{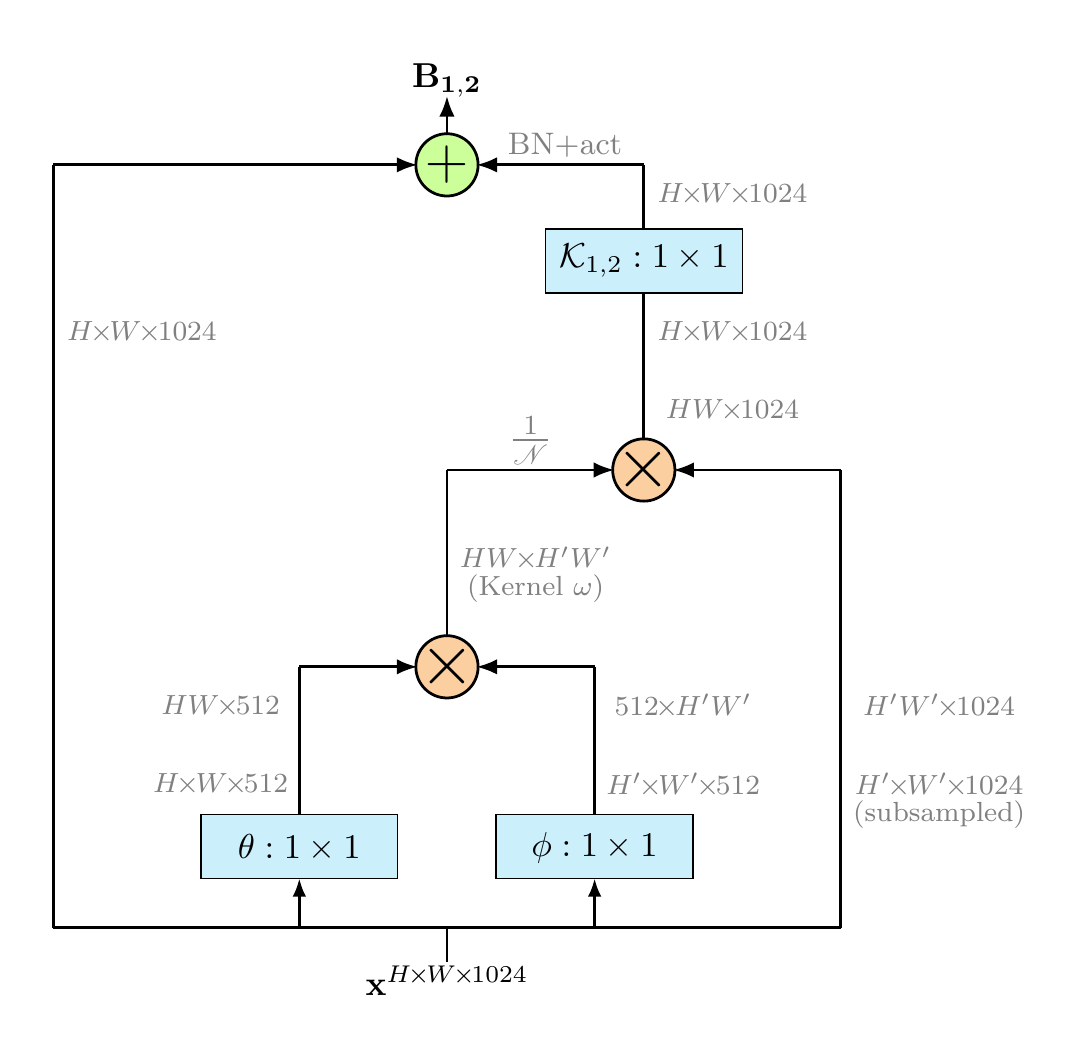}
\caption{Forward propagation of tensors in a \emph{subsampled} nonlocal block with the inverse Laplacian operator $(-\Delta)^{-s}$.} \label{fig:tensordiag4}
\end{figure}

\subsubsection*{Subsampled nonlocal blocks} 
As we can see in Section 6.1, the global computation can be quite expensive. In order to minimize the floating-point operations, several remedies exist. Firstly, all the convolutions in the nonlocal block ($\theta, \phi, \mathcal K_1, \mathcal K_2$) are $1\times 1$ convolutions. This itself reduces the computational effort drastically. Secondly, the embeddings $\theta$ and $\phi$ can be used to reduce the number of channels by half, as shown in Figure \ref{fig:tensordiag1}. This reduces the computation of a nonlocal block by half when the kernel $\omega$ is computed. 

There is one more way of reducing the computational effort of the nonlocal blocks, namely subsampling the image before the affinity of the pixel strips is computed. This is shown in Figure 2 (right). Figure 2 shows how the input $\mathbf X$ (of shape $H\times W\times C$) is spatially subsampled to $\mathbf{\hat X}$ (of shape $H'\times W'\times C$) before the affinity between the regions of the image is computed. Experiments have shown that this has very little impact on the quality of the results of the network. This tells us that we do not need to compare each pair of pixel strips. Instead, it is good enough if we compare patches of the image with each other and learn the correlations between them while we compute the kernel $\omega$. The computational savings due to this subsampling trick \cite{helocal2} are discussed in Section 6.1.

\section{Implementation details and choice of hyperparameters}\label{sec:add}
All the networks are implemented and trained using the Tensorflow library \cite{tensorflow2} with a single Nvidia Tesla P100 GPU. The networks for the semantic segmentation task (Section 5.3) are trained using an Nvidia Tesla V100 GPU. All the trainable weights in the neural network are initialized based on the suggestions in \cite{init12}. The discretization step size $h$ is kept between $0.04$ and $0.08$. Very small values of $h$ lead to slow propagation of the information down the network, and as a result, the network performs worse. On the other hand, we see exploding gradients and sudden feature transformations in the network for bigger values of $h$, which cause instability and adversely affect training and convergence.

The loss function $S$ in equation (2.2) is chosen to be the cross-entropy loss function. Stochastic Gradient Descent (SGD) is used as the optimizer with momentum 0.9 and with a learning rate of 0.01 for the first epoch to warm up the training. Then we go back to a learning rate of 0.1 and reduce it by a factor of 10 after 80, 120, 160 and 180 training epochs. The weight decay constant $\alpha_1$ for weights in the normal blocks is $2\times 10^{-4}$ for CIFAR-10/CIFAR-100/BDD100K and $5\times 10^{-4}$ for STL-10. The weight smoothness decay constant $\alpha_2$ is $1\times 10^{-8}$. The reason for such a small value for $\alpha_2$ is that, in our case, we have a nonlocal block after the second block in each Unit. While we do want the weights to vary smoothly, the weights and the features of the second and third Hamiltonian block will be invariably quite different because of the presence of the nonlocal operation between the two blocks. Therefore, $\alpha_2$ is kept smaller than the value suggested in \cite{rev2}. As discussed before, the convolutional weights in the nonlocal block are only regularized by weight decay ($2\times10^{-4}$ for all the datasets, i.e. CIFAR-10/100, STL-10, BDD100K) and not by the weight smoothness decay. The scaling factor $\lambda$ is 0.1 for all the nonlocal operators with the dimensional constant $n=2$. The value of $s$ for the pseudo-differential operators is $1/2$. From the experiments, it has been seen that the nature of the nonlocal operator plays a more important role than the power of the fractional Laplacian and the inverse fractional Laplacian. This is to say, any value of $s$ between 0 and 1 works for the two pseudo-differential operators, and the performance is rather determined by the nature of the nonlocal interaction between the pixels strips and not by the value of $s$ in the kernel of the integral operator.

\section{Computational costs for the STL-10 dataset} \label{sec:compstl}
When it comes to datasets like STL-10 \cite{stl2} (or ImageNet, etc.), there is a trade-off that needs to be taken into consideration. Often, the training of networks is bound by memory constraints and not by computational constraints. In such cases, this increase is within acceptable limits. But when the aim is to bring down the number of floating-point operations, one could use the subsampling trick in the nonlocal block (Figure 2 and Section \ref{sec:speedup} in this supplementary material). This drastically reduces the number of floating-point operations for the computations of the kernel $\omega$ while measuring the affinity between each pair of pixel strips/sections of the image, without changing the nonlocal nature of the block. Table \ref{tab:subsample} shows the effect of the subsampling pool size on the number of floating-point operations for the nonlocal diffusion network while training on the STL-10 dataset.\renewcommand{\arraystretch}{1.3}
\begin{table}
\centering
\caption{Number of floating-point operations depending on the subsampling pool size in the nonlocal blocks while training the nonlocal diffusion network on the STL-10 dataset.}
\scalebox{0.8}{%
\begin{tabular}{|l|c|c|}
\hline
\multicolumn{1}{|c|}{\multirow{2}{*}{\textbf{Network}}} & \textbf{Subsample}  & \textbf{STL-10} \\ \cline{3-3} 
\multicolumn{1}{|c|}{} & \textbf{pool size} & \textbf{\footnotesize FLOPs (M)} \\ \hhline{|=|=|=|}
\multirow{6}{*}{Nonlocal diffusion $\mathcal L$}& 0 & 9341.2 \\
 & 2 & 3471.4 \\
 & 4 & 2003.7  \\
 & 6 & 1731.9  \\
 & 8 & 1636.8  \\
 & 12 & 1568.9 \\ \hline
Hamiltonian-74 & NA\footnotemark & 1432.5 \\ \hline
\end{tabular}%
}
\label{tab:subsample}
\end{table}\renewcommand{\arraystretch}{1}
\footnotetext[1]{Not applicable because the original Hamiltonian network has no nonlocal blocks.}The zero in the table stands for non-subsampled nonlocal blocks. Clearly, the computational cost is quite high if no subsampling is performed. When the pool size is increased, the number of floating-point operations gradually approaches the number of FLOPs for the original Hamiltonian network (on STL-10). When the pool size is too large, the nonlocal character of the block is degraded a bit. Thus, it is a balancing act between computational costs versus the increase in performance of a network in the presence of nonlocal blocks. One has to choose the fitting pool size for the subsampling in the nonlocal block, based on the computational constraints that one has and the resolution of each image.
\newpage
\section{Real parts of the eigenvalues of weight matrices $\mathcal K_1$, $\mathcal K_2$}
\label{ap:eigunit}
\text{ }
\begin{figure}[htb!]
\centering
\includegraphics[scale=0.3]{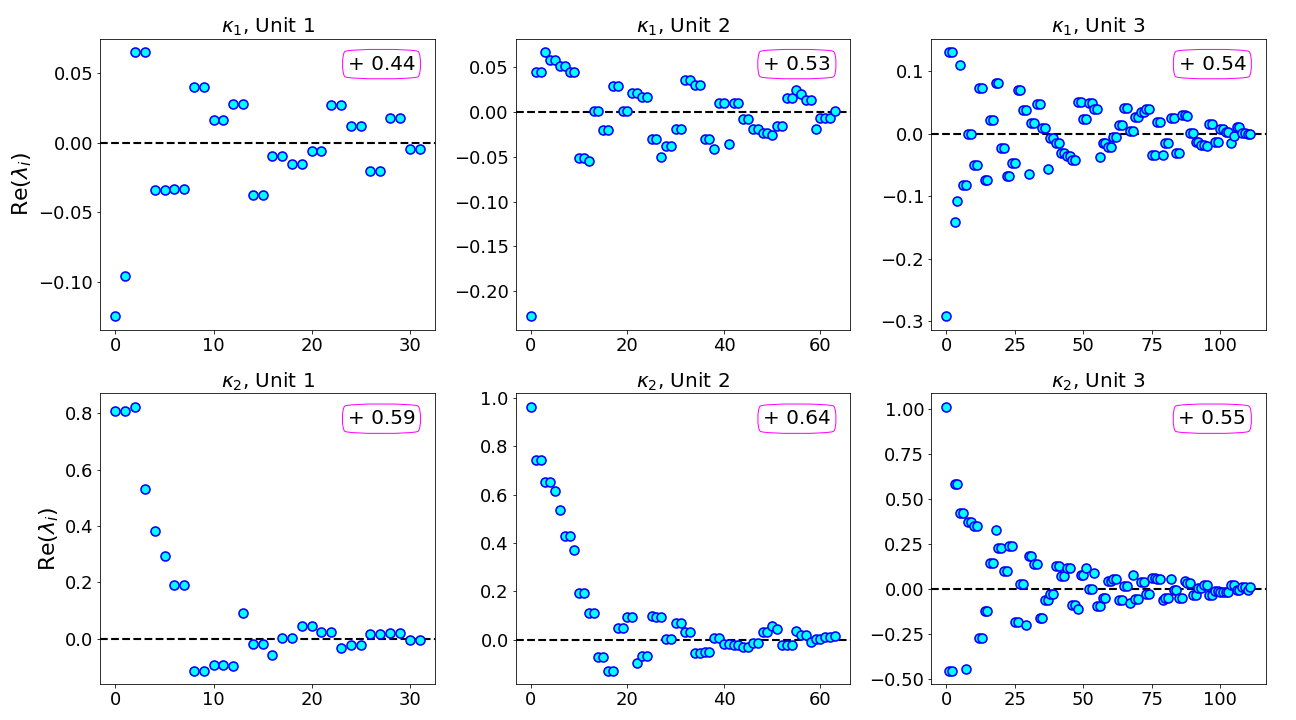}
\caption{Real parts of the eigenvalues of weight matrices $\mathcal K_1$ and $\mathcal K_2$ in the nonlocal block placed in each Unit of the pseudo-differential $(-\Delta)^{1/2}$ network while training on STL-10.}
\label{fig:eigunit4}
\end{figure}
\begin{figure}[htb!]
\centering
\includegraphics[scale=0.3]{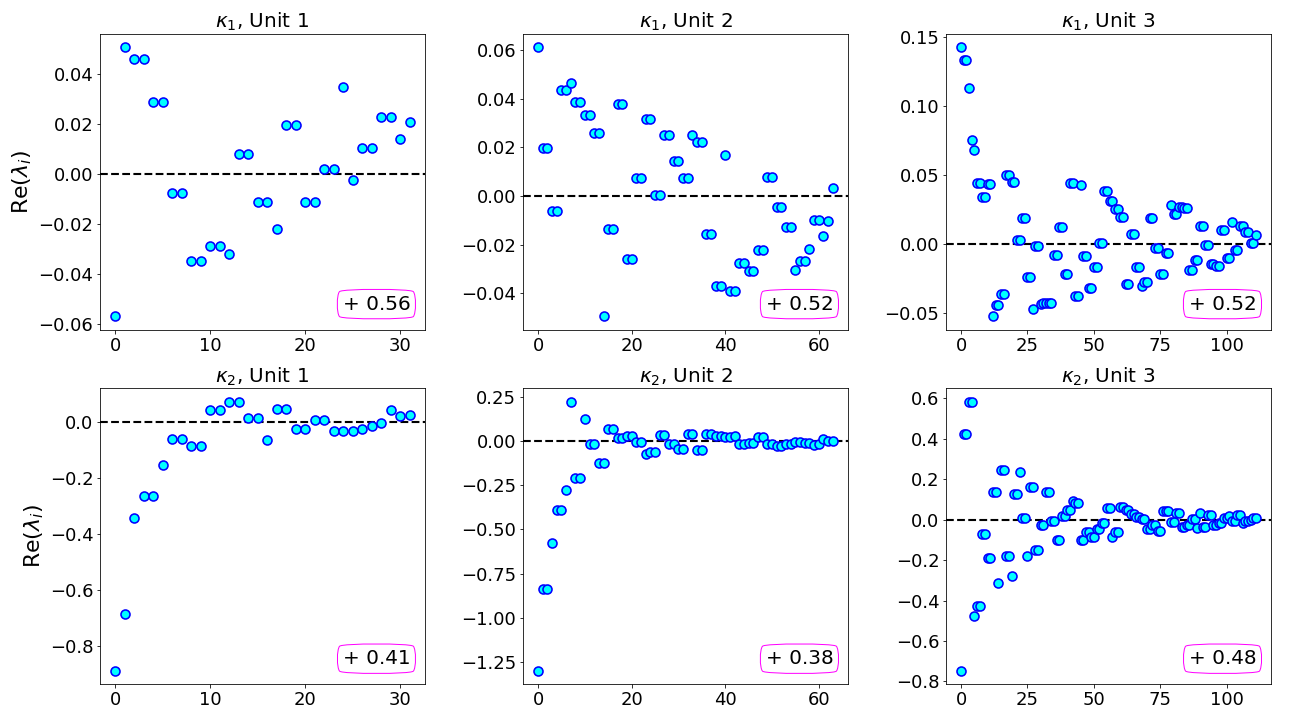}
\caption{Real parts of the eigenvalues of weight matrices $\mathcal K_1$ and $\mathcal K_2$ in the nonlocal block placed in each Unit of the pseudo-differential $(-\Delta)^{-1/2}$ network while training on STL-10.}
\label{fig:eigunit6}
\end{figure}

\clearpage
\begin{figure}[htb!]
\centering
\includegraphics[scale=0.3]{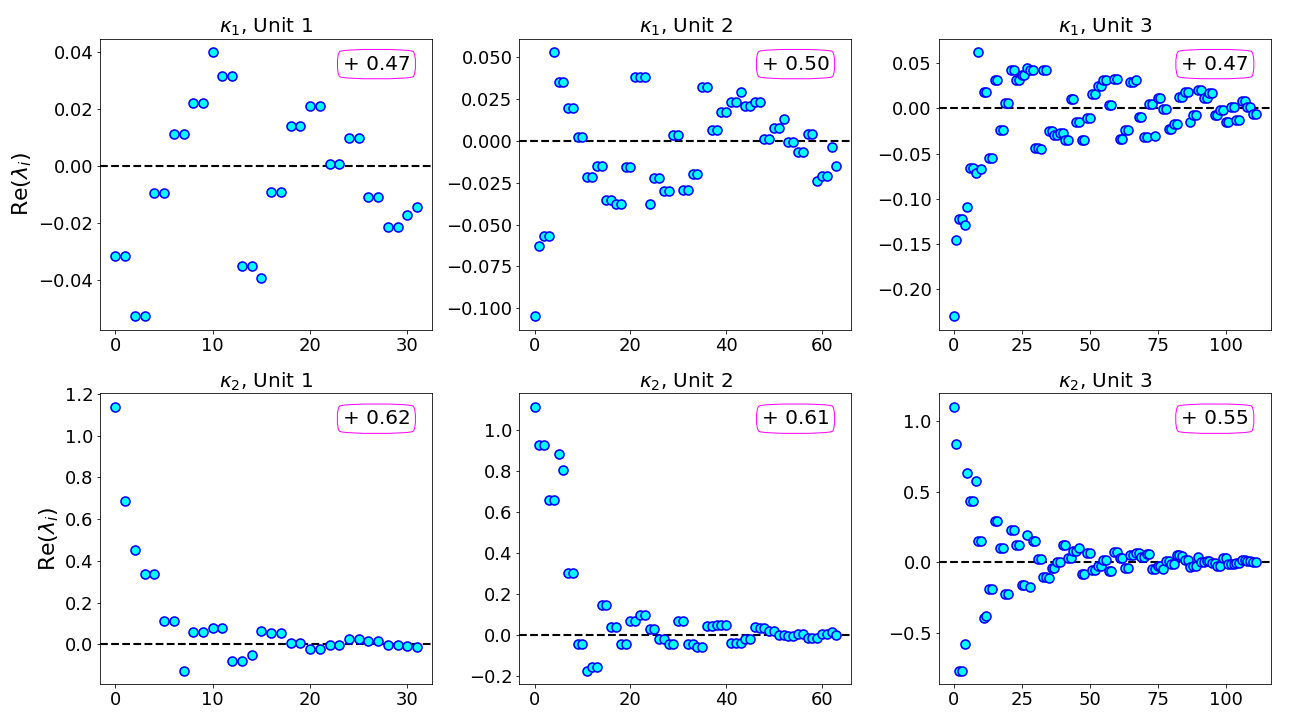}
\caption{Real parts of the eigenvalues of weight matrices $\mathcal K_1$ and $\mathcal K_2$ in the nonlocal block placed in each Unit of the pseudo-differential $(-\Delta)^{-1}$ network while training on STL-10.}
\label{fig:eigunit8}
\end{figure}

\section{Eigenvalues of the symmetric parts of weight matrices $\mathcal K_1$, $\mathcal K_2$}
\label{ap:sym}
\text{ }
\begin{figure}[htb!]
\centering
\includegraphics[scale=0.3]{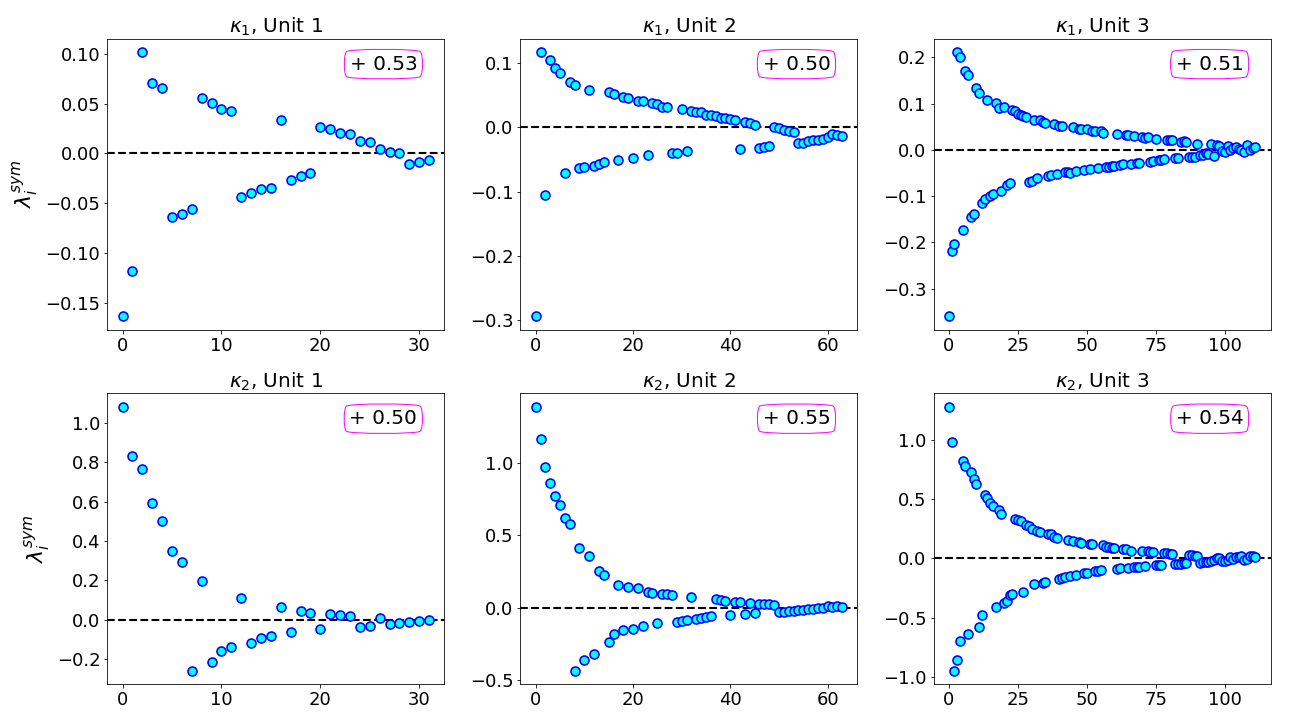}
\caption{Eigenvalues of the symmetric parts of weight matrices $\mathcal K_1$ and $\mathcal K_2$ in the nonlocal block placed in each Unit of the pseudo-differential $(-\Delta)^{1/2}$ network while training on STL-10.}
\label{fig:sym4}
\end{figure}

\begin{figure}[htb!]
\centering
\includegraphics[scale=0.3]{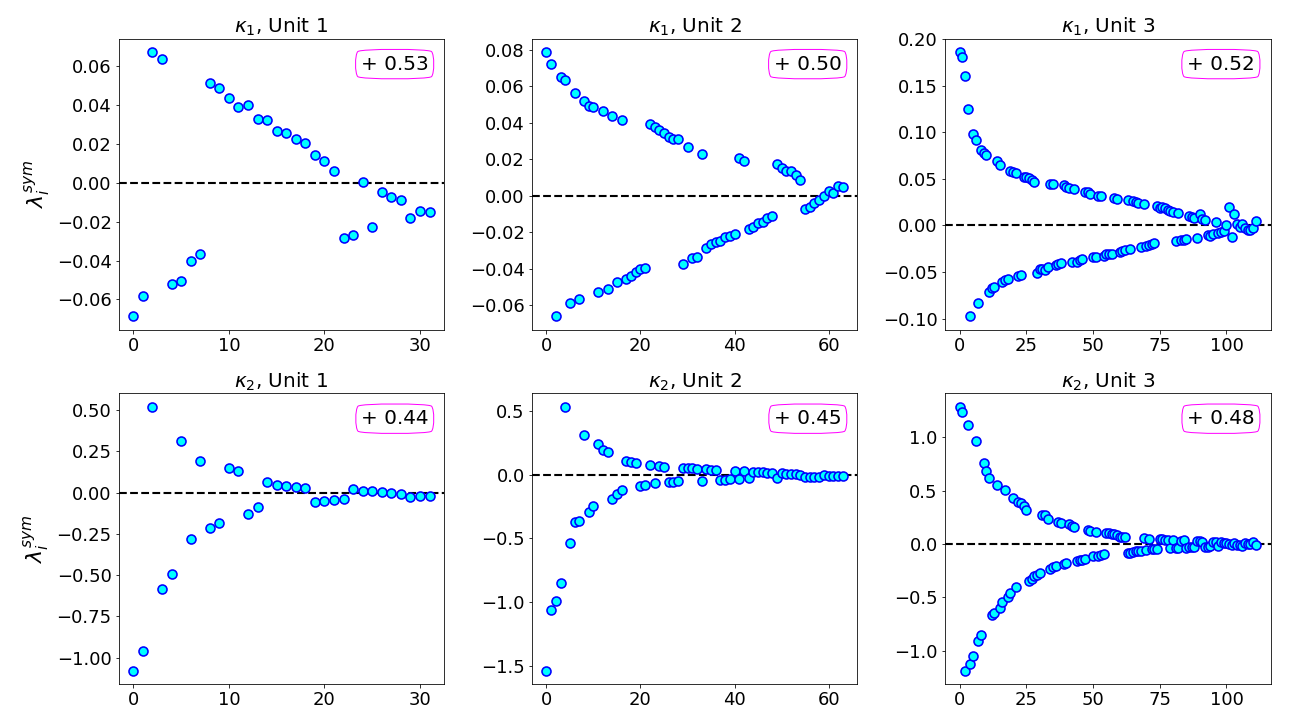}
\caption{Eigenvalues of the symmetric parts of weight matrices $\mathcal K_1$ and $\mathcal K_2$ in the nonlocal block placed in each Unit of the pseudo-differential $(-\Delta)^{-1/2}$ network while training on STL-10.}
\label{fig:sym6}
\end{figure}

\begin{figure}[htb!]
\centering
\includegraphics[scale=0.3]{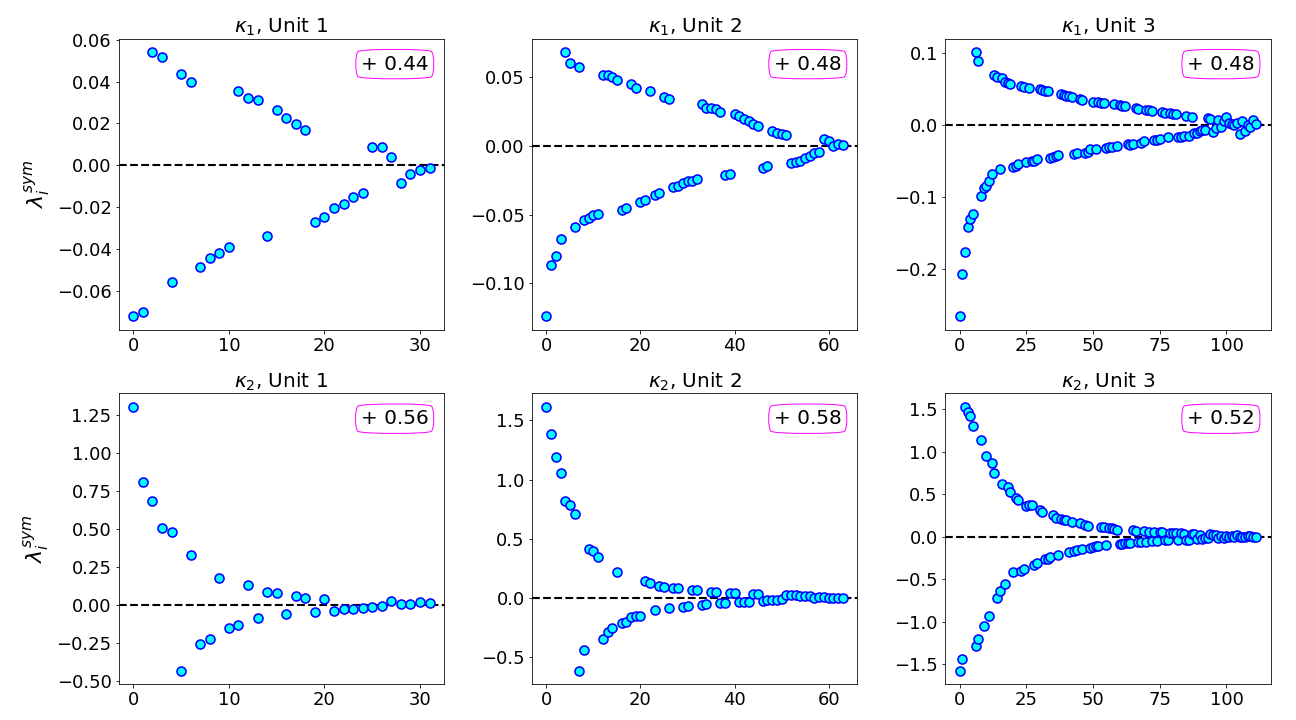}
\caption{Eigenvalues of the symmetric parts of weight matrices $\mathcal K_1$ and $\mathcal K_2$ in the nonlocal block placed in each Unit of the pseudo-differential $(-\Delta)^{-1}$ network while training on STL-10.}
\label{fig:sym8}
\end{figure}

\clearpage
\bibliographystyle{siamplain}
\bibliography{references2}